\documentclass[a4paper,fleqn]{cas-dc}
\usepackage[authoryear]{natbib}
\usepackage{enumitem}
\usepackage{ragged2e} 
\usepackage{amsmath}
\usepackage{algorithm}
\usepackage{algpseudocode}
\usepackage{amsfonts}
\usepackage{multicol} 
\usepackage{caption}
\usepackage{booktabs}
\usepackage{graphicx}
\usepackage{subcaption}
\usepackage{hyperref}

\def\bng{\bngx}

\font\bngx=bang10

\def\*#1*#2{o\null{#2}{#1}}

\def\sh#1{\setbox0=\hbox{#1}%
     \kern-.02em\copy0\kern-\wd0
     \kern.04em\copy0\kern-\wd0
     \kern-.02em\raise.0433em\box0 }
\def\tsc#1{\csdef{#1}{\textsc{\lowercase{#1}}\xspace}}
\tsc{WGM}
\tsc{QE}
\tsc{EP}
\tsc{PMS}
\tsc{BEC}
\tsc{DE}
\newcommand{\customunderline}[2][1.5mm]{%
  \underline{\vphantom{y}\vspace{-#1}\hspace{0pt}#2}%
}

\begin{document}
\let\WriteBookmarks\relax
\def\floatpagepagefraction{1}
\def\textpagefraction{.001}
\shortauthors{\textit{F.T.J. Faria et~al.}}
\shorttitle{Vashantor: A Large-scale Multilingual Benchmark Dataset for Automated Translation of Bangla Regional Dialects to Bangla Language}

\title[mode=title]{Vashantor: A Large-scale Multilingual Benchmark Dataset for Automated Translation of Bangla Regional Dialects to Bangla Language}

\author[1]{Fatema Tuj Johora Faria} [orcid=0009-0005-1684-8211]

\address[1]{Department of Computer Science and Engineering, Ahsanullah University of Science and Technology, Dhaka 1208, Bangladesh}
\ead{fatema.faria142@gmail.com}

\author[1]{Mukaffi Bin Moin} [orcid=0009-0001-9634-5809]

\ead{mukaffi28@gmail.com}
\author[1]{Ahmed Al Wase} [orcid=0009-0000-2553-4891]
\ead{ahmed.alwasi34@gmail.com}

\author[2]{Mehidi Ahmmed}
\ead{mehidi0308@gmail.com}
\address[2]{Environmental Sciences, Khulna University, Khulna, 9208, Bangladesh}

\author[1]{Md. Rabius Sani}
\ead{rshridoy010113@gmail.com}

\author[3]{Tashreef Muhammad} [orcid=0000-0002-9816-2547]
\address[3]{Department of Computer Science and Engineering, Southeast University, Dhaka 1208, Bangladesh}
\ead{tashreef.muhammad@seu.edu.bd}
\cormark[1]

\cortext[cor1]{Corresponding author}

\begin{abstract}
The Bangla linguistic variety is a fascinating mix of regional dialects that contributes to the cultural diversity of the Bangla-speaking community. Despite extensive study into translating Bangla to English, English to Bangla, and Banglish to Bangla in the past, there has been a noticeable gap in translating Bangla regional dialects into standard Bangla. In this study, we set out to fill this gap by creating a collection of \textbf{32,500} sentences, encompassing Bangla, Banglish, and English, representing five regional Bangla dialects. Our aim is to translate these regional dialects into standard Bangla and detect regions accurately. To tackle the translation and region detection tasks, we propose two novel models: \textbf{DialectBanglaT5} for translating regional dialects into standard Bangla and \textbf{DialectBanglaBERT} for identifying the dialect’s region of origin. DialectBanglaT5 demonstrates superior performance across all dialects, achieving the highest BLEU score of 71.93, METEOR of 0.8503, and the lowest WER of 0.1470 and CER of 0.0791 on the Mymensingh dialect. It also achieves strong ROUGE scores across all dialects, indicating both accuracy and fluency in capturing dialectal nuances. In parallel, DialectBanglaBERT achieves an overall region classification accuracy of 89.02\%, with notable F1-scores of 0.9241 for Chittagong and 0.8736 for Mymensingh, confirming its effectiveness in handling regional linguistic variation. This is the first large-scale investigation focused on Bangla regional dialect translation and region detection. Our proposed models highlight the potential of dialect-specific modeling and set a new benchmark for future research in low-resource and dialect-rich language settings.
\end{abstract}


\maketitle

\section{Introduction}
\label{introlab}
Neural Machine Translation (NMT) \cite{intro1} represents an innovative technology in Natural Language Processing (NLP),
 bringing about a significant transformation in the way we approach automated translation tasks. Unlike traditional rule-based or statistical machine translation systems, NMT relies on deep neural networks to directly translate text from 
 one language to another. Translation, the core function of NMT \cite{seq}, can be categorized into two primary types: sentence-level translation and word-level translation. Sentence translation involves translating entire sentences or phrases from one language to another, preserving the meaning and context. Word translation, on the other hand, focuses on individual words or short phrases and their corresponding translations. These two types of translation serve distinct purposes \cite{intro3}, with sentence translation allowing comprehensive document translation, w\-hile word translation supports finer-grained language analysis and understanding. While a few years ago, the focus was on 
 achieving high-quality translations for widely spoken and well-resourced languages, the current improvements in translation quality have highlighted the importance of addressing low-resource languages and dealing with more diverse and remarkable translation challenges \cite{TNMT}. 

The Bangla language is spoken by about 228 million people as their first language and an additional 37 million people speak it as a second language. Bangla is the fifth most spoken first language and the seventh most spoken language overall in the world \cite{intro5}. Bangladesh has 55 regional languages spoken in its 64 districts, while the majority of the population speaks two different varieties of Bengali. Some people also speak the language of the region they live in. The variations in the Bengali language extend beyond vocabulary to differences in pronunciation, intonation, and even grammar. A regional language, which is also called a dialect, is a language that children naturally learn without formal grammar lessons, and it can differ from one place to another. These regional languages can cause changes in the way the main language sounds or is written. Even though there are these regional differences, the Bangla language in Bangladesh can be categorized into six main classes: Bangla, Manbhumi, Varendri, Rachi, Rangpuri, and Sundarbani \cite{intro6}.
Our research addresses a significant gap in the field of machine translation, specifically about the Bengali language. While previous research has mostly concentrated on translating between Bengali to English \cite{seq}, 
English to Bengali \cite{intro5},
English to Banglish \cite{intro7}, 
Manipuri to Bengali \cite{intro8}, 
Bangla to Banglish \cite{intro9}, 
and even Hindi to Bangla \cite{intro10}, 
it has largely skipped over translating various Bengali regional dialects to and from Bengali. Furthermore, previous research concentrated mostly on distinguishing regions from audio speech \cite{intro6}, \cite{Audio2}  rather than giving complete translations of regional speeches into Bengali. The overall lack of prior work or dedicated datasets for regional dialect translation emphasizes the importance of our research in filling these significant knowledge gaps. While a variety of dialects is a source of cultural pride, it also poses an enormous communication challenge. Multiple dialects existing can lead to misconceptions, miscommunications, and a communication gap between individuals from different regions. In such a situation, the need for effective machine translation capable of overcoming dialectal differences becomes critical. Our research tackles this critical problem by presenting the ``Vashantor'' dataset, a useful resource for automated translation of Bangla regional dialects to Bangla Language, enabling enhanced communication and understanding across all of these different geographic regions.  

Recognizing Bangladesh's linguistic diversity, we have chosen to concentrate on translating the Bengali language from five distinct regional dialects found in the Chittagong, Noakhali, Sylhet, Barishal, and Mymensingh regions. For example, when we translate the Chittagong regional dialect,  {\bng``bUtidn phr etNNayaer ed{I}lam''} into Bengali, it becomes  {\bng``Aenk idn pr etamaek edkhlam''}. Similarly, when we translate the Noakhali regional dialect, {\bng``emla idn Hr etamar edya Ha{I}lam''} into Bengali, it also turns into {\bng``Aenk idn pr etamaek edkhlam''}. Likewise, the Sylhet regional dialect, {\bng``bak/kaidn baed tumaer edkhlam''} is translated as {\bng``Aenk idn pr etamaek edkhlam''} in Bengali. These translations show how different regional dialects can be expressed in the Bengali language. These regions have been identified as the most significant contributors to the research due to the pronounced linguistic variations and communication challenges presented by their unique dialects. To the best of our knowledge, this is the first attempt to translate Bangla regional dialects into Bangla. \newline 
In our research, we proposed the DialectBanglaT5 model to facilitate the translation of various regional dialects into standard Bangla. We compared its performance with baseline multilingual and Bangla-specific translation models, such as mT5 and BanglaT5. Additionally, we introduced another model, DialectBanglaBERT, designed to identify the specific regions associated with the text in our corpus. This model was evaluated against existing baseline text classification models, including BanglaBERT and mBERT. In essence, our work has made several noteworthy contributions, which can be summarized as follows:

\begin{itemize}

    \item We created a comprehensive dataset, including:
        \begin{itemize}
            \item 2,500 samples for Bangla, Banglish, and English each.
            \item 12,500 samples for regional Bangla dialects and regional Banglish dialects each.
            \item 12,500 samples for region detection.
        \end{itemize}
    \item We validated the dataset using Cohen's Kappa and Flei\-ss's Kappa.
    \item We applied cosine similarity to quantitatively assess variations and similarities among Bangla regional dialects and the standard Bangla language. 
\item We proposed two novel models: \textbf{DialectBanglaT5}, designed for translating regional dialects into standard Bangla, and \textbf{DialectBanglaBERT}, aimed at identifying the specific regions associated with dialectal text.

    \item We employed machine translation evaluation metrics for assessing the quality of machine translation output, including:
    Character Error Rate (CER), Word Error Rate (WER), Bilingual Evaluation Understudy (BLEU), Recall-Oriented Understudy for Gisting Evaluation (R\-OUGE), and Metric for Evaluation of Translation with Explicit ORdering (METEOR)
    \item We utilized performance metrics, including Accuracy, Precision, Recall, F1 score, and Log loss for the region detection task. 
    \item In our experiments, DialectBanglaT5 clearly outperformed both mT5 and BanglaT5 on each metrics, and DialectBanglaBERT beat BanglaBERT and mBERT on each metrics. 
\end{itemize}

The remaining part of the paper is structured as follows: Section \ref{Relatedlab} provides a thorough review of related works that serve as the foundation for our study. Moving forward, Section \ref{Corpuslab} delves into the details of our dataset creation process. The Section \ref{Modelslab} explores the complexities of the models used for dialect-to-Bangla translation and region detection. In  Section \ref{EvalLab}, we thoroughly investigate the evaluation metrics used to evaluate both regional dialect translation and region detection. The Section \ref{Methodologylab} presents and goes further on our proposed methodology, demonstrating the analytical process and planning that explains our research goals. The Section \ref{Resultslab} diligently describes the results of the experiment and gives a comprehensive understanding. The Section \ref{Futurelab} focuses further on identifying future research opportunities and providing a road map for the current study's continuation and advancement. The Section \ref{Conclusionlab} brings the research to a conclusion by bringing together the many components of our study and offering a final summary that describes the key results, contributions, and significance of our research.

\section{Related Works} \label{Relatedlab}
In this section, we have given a summary of previous research which is relevant to our research. The summary is broken down into four primary subsections: Section \ref{sec2.1} Unsupervised Neural Machine Translation (UNMT), focusing on translation learning without paired examples, using monolingual data from both languages involved; Section \ref{sec2.2} Sequence-to-Sequence (Seq2Seq) Neural Machine Translation employing an encoder-decoder setup to convert text from one language to another; Section \ref{sec2.3} Transformer Based Neural Machine Translation utilizing self-attention to capture word dependencies for improved translation performance and efficiency; and Section \ref{sec2.4} Adversarial Neural Machine Translation, integrating adversarial learning techniques to differentiate between human and machine translations, aiming to enhance translation quality. Table \ref{tab:mytable} and \ref{tab:mytable1} present a comprehensive summary of the existing works on machine translation that have been discussed within this study.
\begin{table*}

\caption{A summary of various existing unsupervised, sequence-to-sequence (Seq2Seq) neural machine translation works }  \label{tab:mytable}

  \begin{tabular}{p{3.5cm} p{4.0cm} p{6.5cm}}

    \toprule 
    Types & Authors  & Contribution  \\
\midrule
     &   Lenin et al. \cite{manipuri}  & Used unsupervised Machine Translation models for the low-resource Manipuri-English language pair.   \\ \cmidrule{2-3}
 \textbf{Unsupervised \newline Neural  Machine Translation} &  Guillaume et al. \cite{UNMT2}  & Explored machine translation in the absence of parallel data and put out a strategy for aligning monolingual corpora.\\ \cmidrule{2-3}
&  Zhen et al. \cite{rel1} & Utilized an extension to extract high-level representations of the input phrases, which consists of two independent encoders sharing partial weights.\\ \midrule

       &   Rafiqul et al. \cite{seq}  & Focused on Bengali to English translation, addressing Bengali's complex syntax and rich vocabulary. They achieved under 2\% loss by merging Seq2Seq and attention-based RNNs using cross-entropy metrics.\\ \cmidrule{2-3}
  \textbf{Sequence-to-Sequence (Seq2Seq) Neural Machine Translation} &  Shaykh et al. \cite{intro5}   & Concentrated on English to Bangla translation employing an encoder-decoder RNN structure with a knowledge-based context vector for precise translation.\\ \cmidrule{2-3}

&   Arid et al. \cite{intro1} & Investigated several neural machine translation techniques for Bangla-English, with average improvements of 14.63\% and 32.18\%.\\ \cmidrule{2-3}
&   Khered et al. \cite{seq3} & verified Levantine and Gulf dialects in the Dial2MSA dataset using human annotators and introduced Dial2MSA-Verified and benchmarked Seq2Seq models, with AraT5 achieving the highest average BLEU score of 41.12. \\
     \bottomrule
    
  \end{tabular}
\end{table*}
\begin{table*}
\caption{A summary of various existing  transformer-based, adversarial network-based neural machine translation works}\label{tab:mytable1}
    \begin{tabular}{p{3.5cm} p{4.0cm}  p{6.5cm}}
    \toprule
    Types & Authors  & Contribution  \\
\midrule
     &  Laith et al. \cite{TNMT}  & Presented a Transformer-Based NMT model intended for Arabic dialects, solving issues in low-resource languages through the use of subword units.  \\ \cmidrule{2-3}
 \textbf{Transformer-Based Neural Machine Translation}  &  Soran et al. \cite{TNMT2} & Suggested a unique transformer-based NMT model adapted for the low-resource Kurdish Sorani Dialect, earning an impressive BLEU score of 0.45 for high-quality translations.  \\ \cmidrule{2-3}
&  Dongxing et al. \cite{rel3}  & Presented the interacting-head attention method, which improves multihead attention by allowing more extensive and deeper interactions among tokens in different subspaces. \\ \cmidrule{2-3}
&  Li  et al. \cite{TNMT3}  & introduced JLMS25 and a knowledge transfer method, cutting CER and WER by up to 7.7\% and 10.8\% for Jiao-Liao Mandarin. \\ 
\midrule
    &   Lijun et al. \cite{ANMT}  & Developed Adversarial-NMT technique with a CNN to minimize differences between human and NMT translations.  \\ \cmidrule{2-3}
\textbf{Adversarial \newline Neural Machine Translation}   & Wenting et al. \cite{ANMT2}  & Utilized a GAN for Chinese-English translation which surpassed RNNs by 8\% in translation quality on an English-Chinese dataset and achieved an average BLEU score of 28.2.  \\ \cmidrule{2-3}
 & Wei et al. \cite{rel4} & Explored a reinforcement learning paradigm that includes a discriminator as the terminal signal to limit semantics.
\\       \bottomrule
    
  \end{tabular}
\end{table*}
\subsection{Unsupervised Neural Machine Translation} \label{sec2.1}
Lenin et al.  \cite{manipuri} introduced unsupervised Machine Translation models for low-resource languages, emphasizing their ability to work without parallel sentences. It focused on Manipuri-English, highlighting linguistic disparities and challenges. It concluded that unsupervised Machine Translation for such language pairs is feasible, based on experiments. They compared Unsupervised Machine Translation (USMT) and UNMT models, focusing on models like Monoses, MASS, and XLM. Initial findings showed that USMT was more effective for Manipuri-English. It highlighted challenges in adapting unsupervised MT methods to this pair and evaluated the strengths and weaknesses of USMT and UNMT models. It used a Manipuri-English corpus from newspapers and evaluated using BLEU scores. Monoses obtained the best BLEU score of 3.13 for English to Manipuri and a score of 6.37 for Manipuri-English outperforming the UNMT systems. They established a baseline and encouraged further research for the low-resource Manipuri-English language pair. In another research, Guillaume et al. \cite{UNMT2} explored the possibility of machine translation in the absence of parallel data, which is a significant challenge in the field. They offered a model for mapping monolingual corpus from two distinct languages into a common latent space. They demonstrated their approach's strong performance in unsupervised machine translation. While it might not outperform supervised approaches with abundant resources, it produced outstanding results. It matched the quality of a supervised system trained on 100,000 sentence pairs from the WMT dataset, for example. It achieved strong BLEU scores in the Multi30K-Task1 dataset, particularly 32.76 in the English-French pair. The research also examined the model's performance in various settings, demonstrating its versatility. The primary outcome was establishing the practicality of unsupervised machine translation using shared latent representations, with outstanding results across a wide range of language pairs.

\subsection{Sequence-to-Sequence (Seq2Seq) Neural Machine Translation} \label{sec2.2}
The work of Rafiqul et al. \cite{seq} focused on translating Bengali to English, which overcomes the difficulties of Bangl\-a's complicated grammatical rules and large vocabulary. To improve performance, they employed a Seq2Seq learning model with attention-based recurrent neural networks (RNN) and cross-entropy loss metrics. They built a model with less than 2\% loss by carefully building a dataset with over 6,000 Bangla-English Seq2Seq sentence pairs and precisely analyzing training parameters. Shaykh et al. \cite{intro5} on the contrary, focused on the task of English to Bangla translation using RNN, especially an encoder-decoder RNN architecture. Their method included a knowledge-based context vector to aid in exact translation between English and Bangla. The study highlighted the importance of data quality, with 4,000 parallel sentences serving as the foundation. Particularly, they overcame the issue of varying sentence lengths by using a combination of linear activation in the encoder and tanh activation in the decoder to achieve optimal results. In addition, their findings highlighted the superiority of GRU over LSTM as well as the significance of attention processes implemented via softmax and sigmoid activation functions. On the other hand, Khered et al. \cite{seq3} extended the Dial2MSA dataset by verifying translations for Levantine and Gulf dialects using human annotators, and benchmarked multiple Seq2Seq models with reporting average BLEU and chrF++ improvements of 41.12 and 62.05 respectively, with AraT5 achieving top performance, especially on the Gulf dialect. 

\begin{figure*}
\centering
\includegraphics[width=1.0\textwidth]{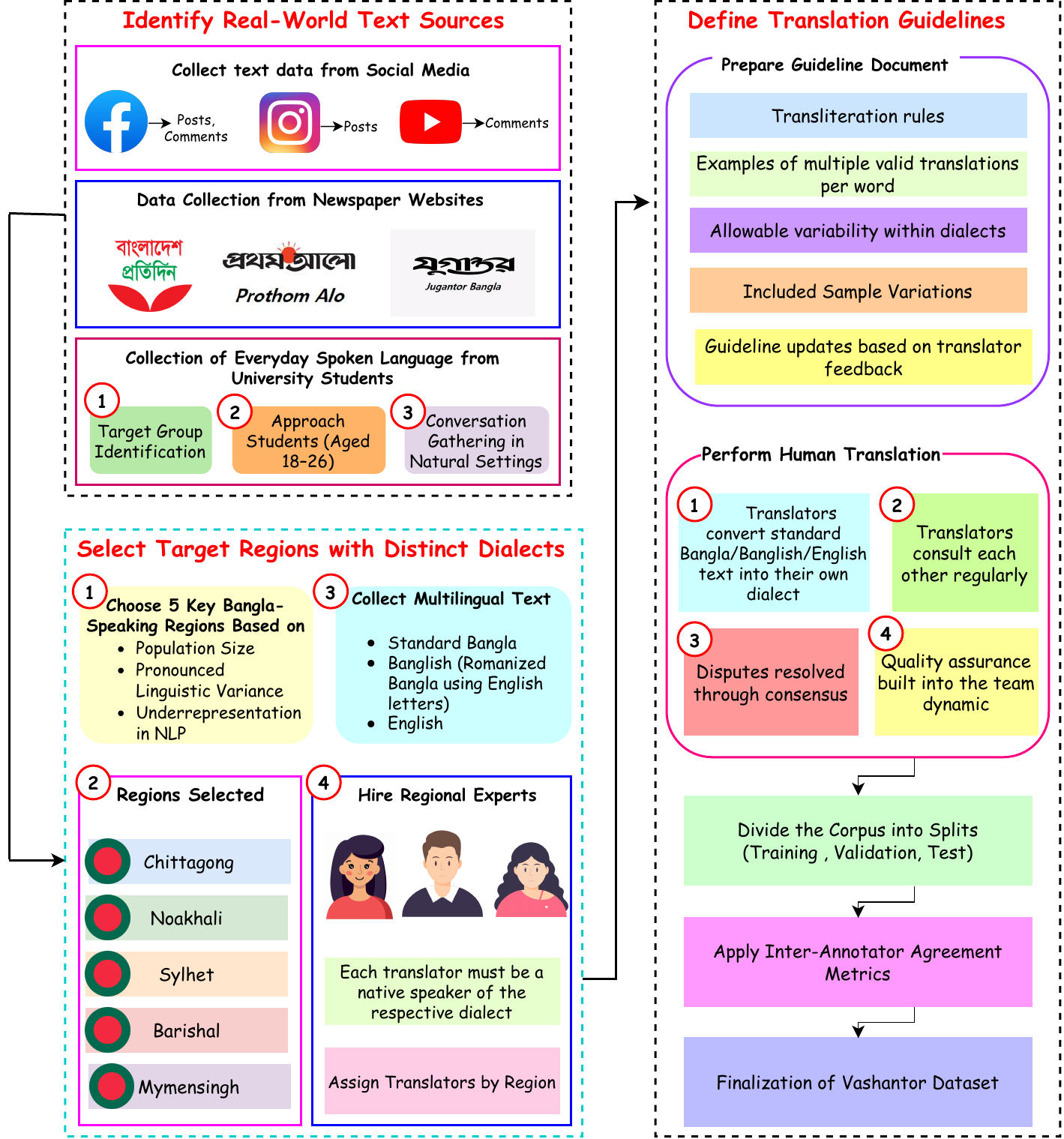}
\caption{Detailed Workflow Illustrating the Construction Process of the Vashantor Dialectal Dataset}\label{fig:Dataset}

\end{figure*}

\subsection{Transformer-Based Neural Machine Translation} \label{sec2.3}
Laith et al. \cite{TNMT} proposed an innovative Transformer-Based NMT model tailored for Arabic dialects, addressing challenges in low resource languages, particularly their unique word order and scarce vocabulary. Their approach employed subword units and a common vocabulary, as well as the WordPiece Model (WPM) for exact word segmentation, sparsity reduction, and translation quality enhancement, particularly for unknown (UNK) terms. Key contributions included a shared vocabulary approach between the encoder and decoder, as well as the usage of wordpieces, which resulted in higher BLEU scores. The research indicated a considerable improvement in translation quality through comprehensive testing including diverse Arabic dialects and translation jobs to Modern Standard Arabic (MSA). Furthermore, the study examined the impact of characteristics such as the number of heads in self-attention sublayers and the layers in encoding and decoding subnetworks on the model's performance. On the contrary, Soran et al. \cite{TNMT2}, provided a novel transformer-based NMT model for low-resource languages, with an emphasis on the Kurdish Sorani Dialect. This model employed attention approaches and data from several sources to get a BLEU score of 0.45, suggesting high-quality translations. The addition of four parallel datasets, Tanzil, TED Talks, Kurdish WordNet, and Auta, expanded the system's domain adaptability. Because of its six-layer encoder and decoder architecture, which was improved by multi-head attention, the model offered excellent translation capabilities. Li et al. \cite{TNMT3} introduced JLMS25, a Jiao-Liao Mandarin corpus, and proposed a multi-dialect knowledge transfer framework that reduced CER and WER by up to 7.7

\subsection{Adversarial Neural Machine Translation} \label{sec2.4}

Lijun et al. \cite{ANMT} introduced a novel approach called Adversarial NMT for NMT. Adversarial NMT reduces the distinction between human and NMT translations, in contrast to standard methods that attempt to maximize human translation resemblance. It uses an adversarial training architecture with a CNN as the adversary. The NMT model aims to produce high-quality translations to deceive the adversary, and they were co-trained using a policy gradient method. Adversarial-NMT greatly increases translation quality compared to strong baseline models, according to experimental results on English to French and German to English translation tasks. For English to French, they employed the top 30,000 most frequent English and French words, and for German to English, they utilized the top 32,009 most frequent words. Comparing their Adversarial-NMT to the baseline models, it performed a better translation on about 59.4\% of the sentences. Furthermore, Wenting et al. \cite{ANMT2} discussed the issue of short sequence machine translation from Chinese to English by introducing a generative adversarial network (GAN). The GAN consists of a generator and a discriminator, with the generator producing sentences that are indistinguishable from human translations and the discriminator separating these from human-translated sentences. To evaluate and direct the generator, both dynamic discriminators and static BLEU score targets are used during the training phase. When compared to typical recurrent neural network (RNN) models, experimental results on an English-Chinese translation dataset showed a more than 8\% improvement in translation quality. The proposed approaches' average BLEU scores were 28.2.

\begin{figure*}
    \centering
    \begin{subfigure}{0.32\textwidth}
        \includegraphics[width = \linewidth]{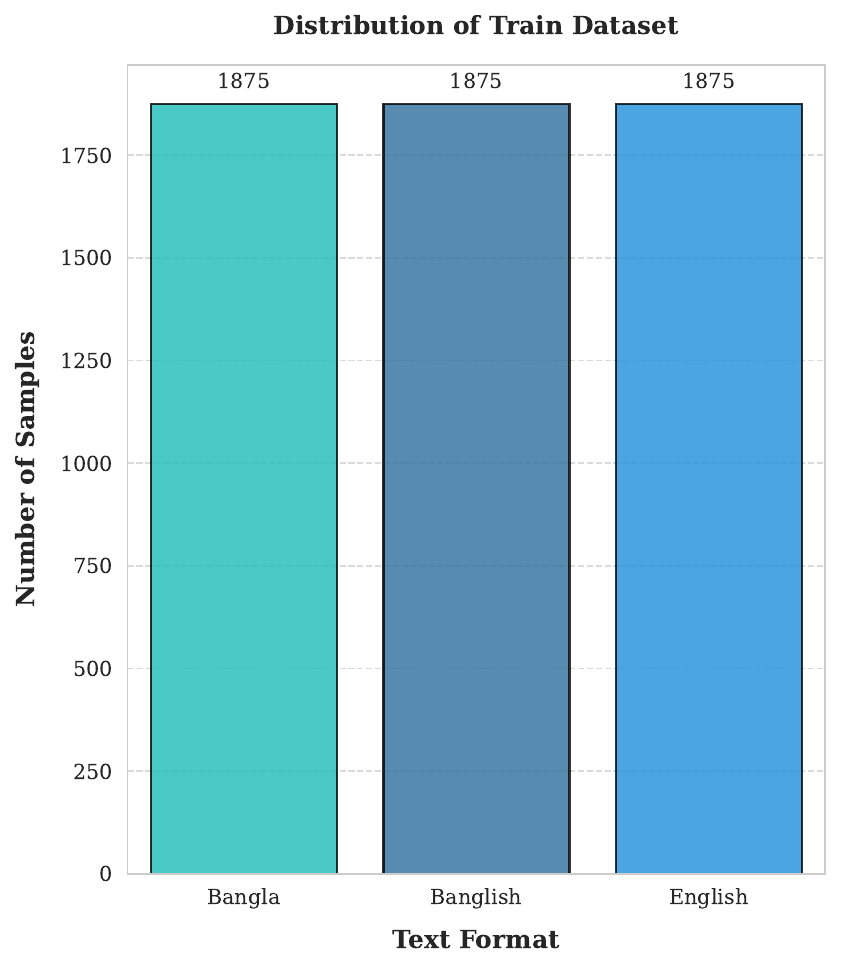}
        
        \caption{Train Dataset}
        \label{fig1}
    \end{subfigure}
    \begin{subfigure}{0.32\textwidth}
        \includegraphics[width = \linewidth]{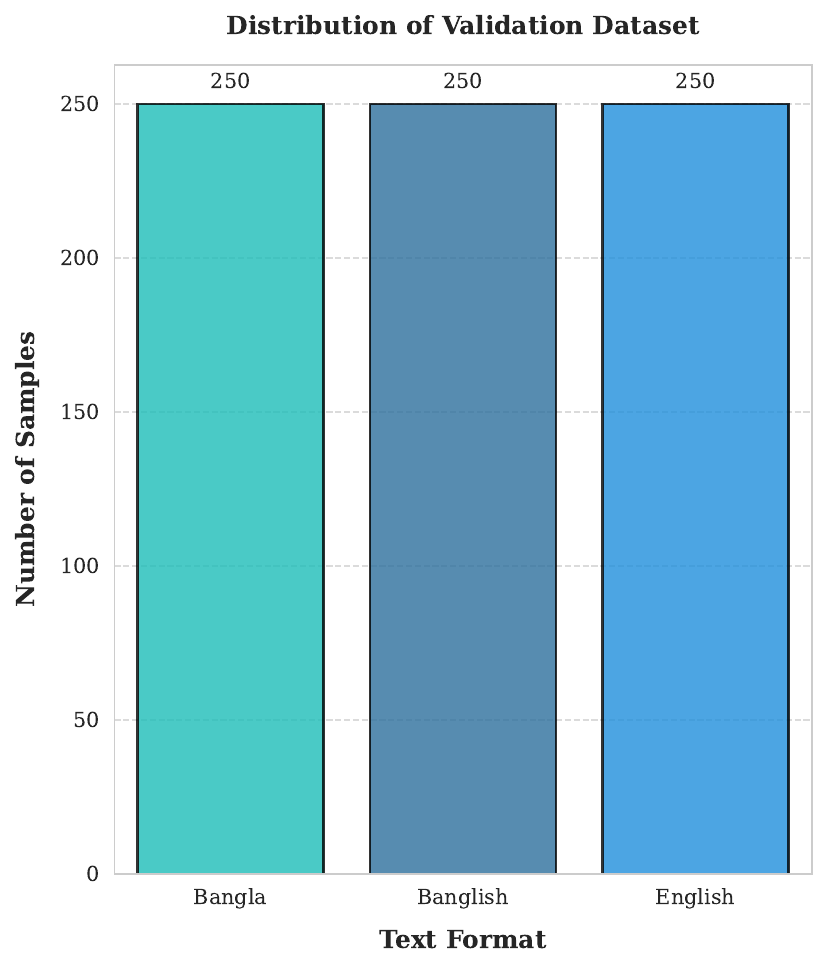}
        
        \caption{Validation Dataset}
        \label{fig2}
    \end{subfigure}
    \begin{subfigure}{0.32\textwidth}
        \includegraphics[width = \linewidth]{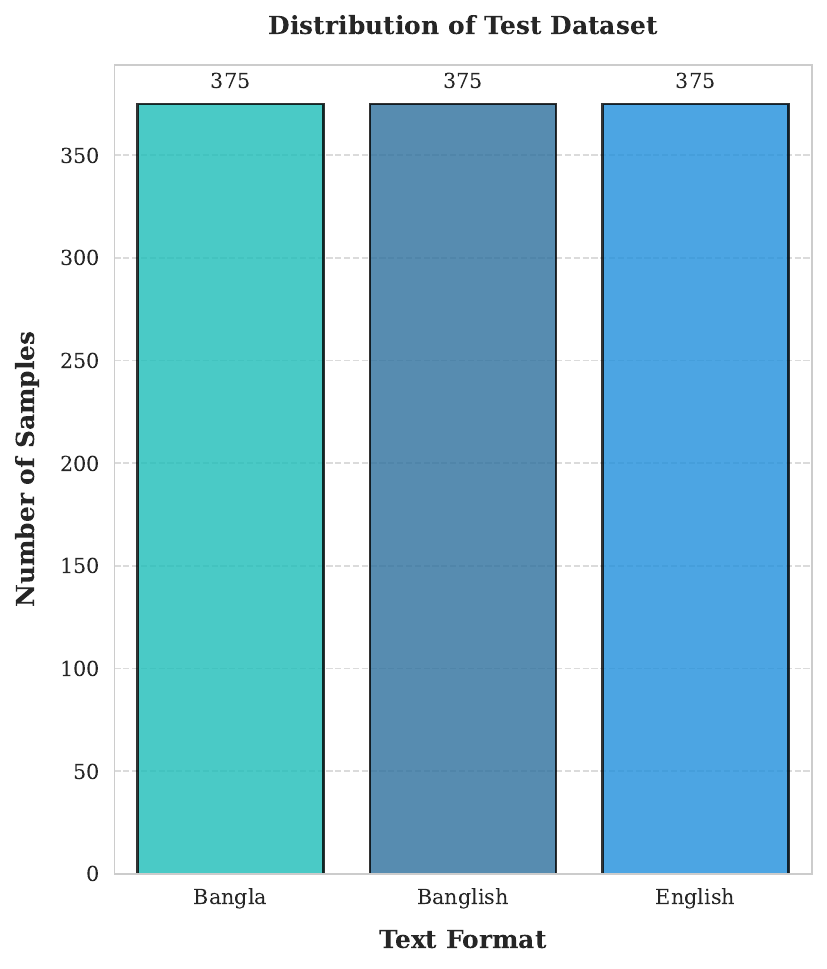}
         \caption{Test Dataset}
        \label{fig3}
    \end{subfigure}
    \caption{\centering Core Data Information}
    \label{fig:core}
\end{figure*}

\begin{table*}
\centering
\caption{Key Demographic and Linguistic Features of Selected Regions in the Vashantor Dataset}
\begin{tabular}{l l l p{5cm}}
\toprule
\textbf{District} & \textbf{Population} & \textbf{Area (km²)} & \textbf{Notes} \\
\midrule
Chittagong & 9,439,076 & 5,282 & Major urban center, dialect hub \\
Noakhali & 3,732,042 & 3,686 & Distinct dialect, rural focus \\
Sylhet & 3,990,003 & 3,452 & Known for unique phonetic traits \\
Barishal & 2,634,203 & 2,785 & Coastal dialect, significant variation \\
Mymensingh & 6,097,814 & 4,395 & Large population, dialect diversity \\
\bottomrule
\end{tabular}
\end{table*}
\begin{figure}
    \centering
    \includegraphics[width=1.0\linewidth]{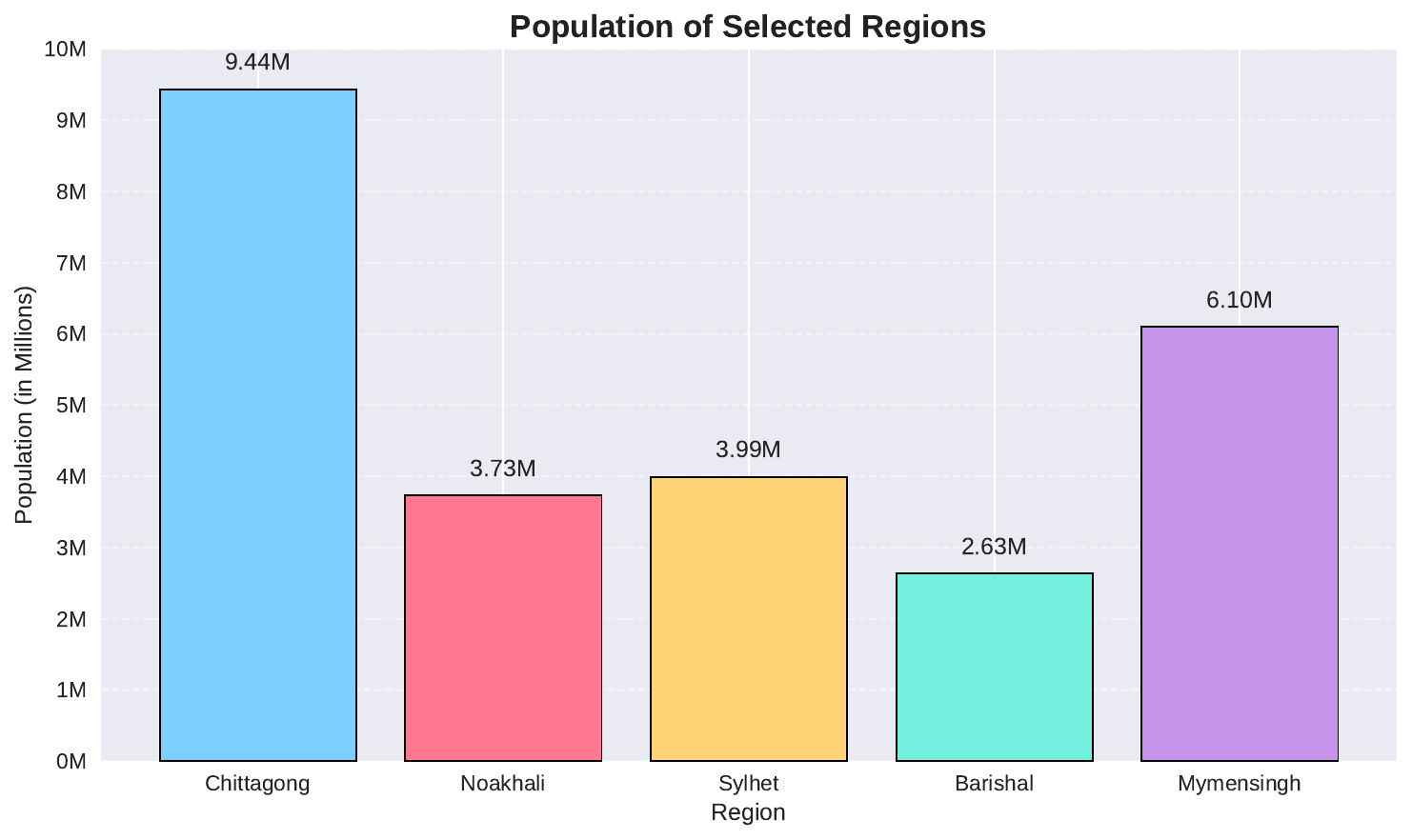}
    \caption{Population Distribution Across Vashantor Dataset Regions}
    \label{fig:enter-label}
\end{figure}
\section{Corpus Creation} \label{Corpuslab}
\subsection{Data Collection}
In the process of curating the ``Vashantor'' dataset, we diligently selected speech text data from a wide range of sources to ensure its authenticity and quality. The name of our dataset was intentionally chosen to be ``Vashantor'' or \textbf{\bng bhaShan/tr} in Bangla, which means ``Translation'' in English. The choice of ``Vashantor'' reflects a deeper cultural connection, especially with regard to the Bangla language itself. It indicates the dataset's focus on Bangla or translations involving Bangla, highlighting the language's significance within the context of the dataset.

We gathered the dataset in Bangla-Banglish, which is a mix of Bangla and English using the English alphabet to write both Bangla and English. It also includes five regional Bangla dialects. Our primary sources for data collection included online platforms, social media, and both formal and informal communication channels. Specifically, we extracted content from Facebook posts and comments, YouTube comment sections, and Instagram comments to capture the natural flow of informal user-generated text. Additionally, to include formal and journalistic language, we collected textual data from major Bangladeshi newspapers such as \textit{Prothom Alo}, \textit{Jugantor}, and \textit{Bangladesh Pratidin}.

To capture everyday spoken language, we also included casual conversations gathered from university students. These chats give us a real glimpse into how young people naturally talk and interact with each other in their daily lives, whether it is during classes, hanging out with friends, or just sharing stories. By prioritizing natural dialogues, discussions, and personal interactions across digital and real-world contexts, we assembled a dataset that mirrors real-life language use. As a result, the ``Vashantor'' dataset is not only diverse and comprehensive but also highly representative of contemporary Bangla language usage in both formal and informal settings.



\subsection{Strategic Region Choices for Bangla Dialect Translation}
The Vashantor dataset targets five Bangla-speaking regions: Chittagong, Noakhali, Sylhet, Barishal, and Mymensingh selected for their distinct dialectal variations in vocabulary, pronunciation, and sentence structure compared to standard Bangla. These regions address communication challenges and the underrepresentation of dialects in NLP, supporting linguistic diversity preservation. Chittagong (9.4 million) is a major dialect hub. Noakhali (3.7 million) and Barishal (2.6 million) offer rural and coastal dialectal diversity, respectively. Sylhet (4.0 million) is chosen for its unique phonetics, and Mymensingh (6.1 million) for its dialectal variation. \cite{bbs2022census} These regions were prioritized for their large populations and linguistic distinctiveness, optimizing the dataset for models like DialectBanglaT5 and DialectBanglaBERT. Other regions were excluded due to resource constraints and lesser dialectal divergence.


\subsection{Translation Process}
In the translation process, we engaged individuals with expertise in each of the five regions, ensuring that the translations were both accurate and consistent. For the Chittagong, Noakhali, and Barishal dialects, three individuals were involved in the translation process. Two translators worked on the Sylhet and Mymensingh regions. Each person played a vital role in understanding the variations of their respective dialects, using their linguistic expertise to translate the text effectively. The translation process was conducted cooperatively, with regular consultations to maintain accuracy and consistency across the dataset. This approach allowed us to capture the distinct features of each dialect while ensuring the dataset's overall quality and reliability.
\begin{table*}
\caption{Translator Information For Bangla Regional Dialects}\label{translatorIdentity}
  \begin{tabular}{llllll} 
    \toprule
    \textbf{Region} & \textbf{Translator} & \textbf{Educational Status} & \textbf{Language Expertise} & \textbf{Age} & \textbf{Gender} \\
    \midrule
    Chittagong & Translator 1 & Undergraduate & Dialect Expert & 25 & Female \\
     & Translator 2 & Undergraduate & Dialect Expert & 24 & Male \\
     & Translator 3 & Graduate & Dialect Expert & 27 & Male \\
     \midrule
   {Noakhali} & Translator 1 & Undergraduate & Dialect Expert & 24 & Male \\
& Translator 2 & Undergraduate & Dialect Expert & 23 & Female \\
& Translator 3 & Graduate & Dialect Proficient & 26 & Male \\ \midrule
    {Sylhet} & Translator 1 & Undergraduate & Dialect Proficient & 25 & Female \\
& Translator 2 & Graduate & Dialect Expert & 27 & Male \\ \midrule
{Barishal} & Translator 1 & Undergraduate & Dialect Expert & 24 & Male \\
& Translator 2 & Undergraduate & Dialect Proficient & 24 & Male \\
& Translator 3 & Undergraduate & Dialect Expert & 25 & Male \\
\midrule
{Mymensingh} & Translator 1 & Undergraduate & Dialect Expert & 23 & Male \\
& Translator 2 & Undergraduate & Dialect Expert & 25 & Male \\
    \bottomrule
  \end{tabular}
\end{table*}
\begin{table*}
\caption{Cosine Similarity Between Bangla Text and Five Bangla Regional Dialects} \label{table2}
\centering
  \begin{tabular}{p{3.0cm} p{2.0cm} p{4.5cm} p{1.6cm}} 
    \toprule
\textbf{Bangla Text} & \textbf{Region Name} & \textbf{Regional Text} & \textbf{Cosine Similarity}  \\
    \midrule
       & Chittagong & {\bng Anr ik phraelHa g{I}reta EebWer{O} gm n laeg?}  & 0.00  \\
       & Noakhali & {\bng Aaen/nr ik Hraelya k{I}et/t Ekechr bhala laeg na?}  & 0.00  \\
     {\bng Aapnar ik prhaelkha kret Ekdm{I} bhal laeg na?} & Sylhet & {\bng Aaphnar ikta prhaelkha kret Ekhdm bhala laeg nain?}  & 0.38  \\
       & Barishal & {\bng Aamenr ik prhalYaHa kret Ek/kael{I} bhaela laeg na?}  & 0.40  \\
       & Mymensingh & {\bng Aapenr ik prhaelHa kret Eek/ker{I} bhala laeg na?}  & 0.86  \\
   
    \bottomrule
  \end{tabular}
\end{table*}
\begin{figure*}
\centering
\includegraphics[width=1.0\textwidth]{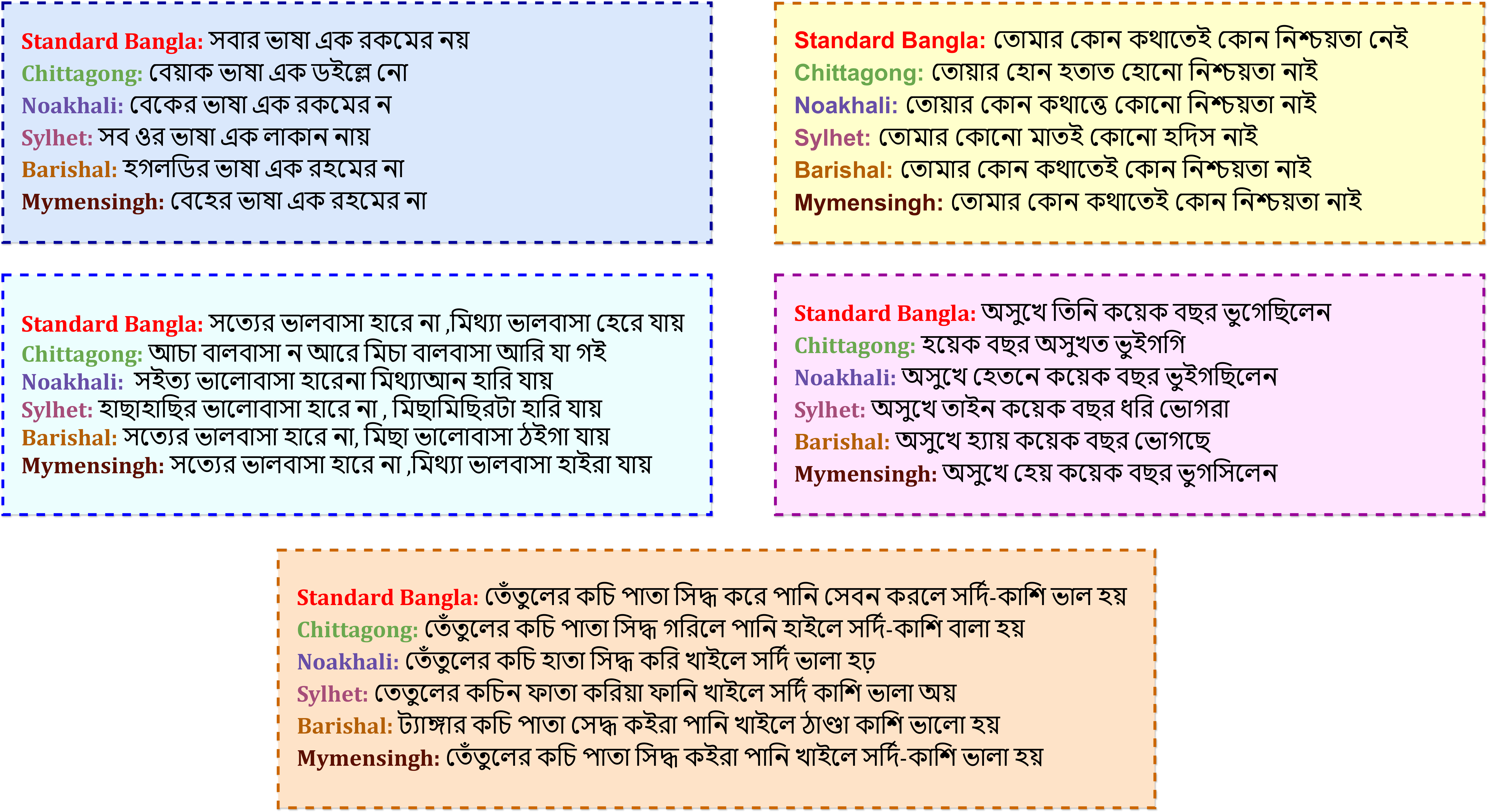}
\caption{Representative Samples from the Vashantor Dataset}\label{fig:Datasample}
\end{figure*}

\subsection{Translation Guideline} 
We provided our translators with guidelines that emphasized authenticity while allowing for regional variability to maintain uniformity in translations. We recognized the value of linguistic variety and incorporated it into our dataset. For example, in the Chittagong region, the word {\bng``tar''} English translation: ``His'' ) can be translated as {\bng``{I}bar''} or {\bng``{I}tar''}. Similarly, {\bng``saeth''} (English translation: ``With") can be expressed as {\bng``l{I}''} or {\bng``ephayaer''}, and {\bng ``gl/p''} (English translation: ``Story") can be written as {\bng``gl/ph''} or {\bng``ikc/ca''}. In Barishal region, multiple words like {\bng``brh bha{I}''} (English translation: ``Elder Brother") conceivably translated as {\bng``nYabha{I}''} or {\bng``emyabha{I}''}. Furthermore, the term {\bng``ebaka''} (English translation: ``fool") is also spoken as {\bng``egaNG/ga''} or {\bng``ebagda''}. Our translation guidelines allowed for different word choices with equivalent meanings, embracing the various writing styles and linguistic diversity found across different regions.
\subsection{Translator Identity}
We engaged with a team of competent and qualified translators, each with specialized expertise in their respective regions, to create the ``Vashantor" dataset. Their qualifications and linguistic competence were essential in assuring the dataset's accuracy and validity. The translators' identities, allocated regions, and linguistic skills are highlighted in Table \ref{translatorIdentity}. \newline

\subsection{Regional Dialect Variations}
Our dataset covers a wide range of dialects and regional differences, showcasing the linguistic diversity across five distinct regions. We used cosine similarity \cite{cosinesimilarity}, a measure of linguistic similarity between two spoken languages, to assess the relationship between Bangla and these regional dialects. This allowed us to quantify linguistic differences and similarities between dialects. For instance, the cosine similarity between the standard Bangla sentence {\bng``Aapnar ik prhaelkha kret Ekdm{I} bhal laeg na?''} (English translation: ``Do you not like to study at all?") and The presentation of equivalents in several dialects in Table \ref{table2} shows how these linguistic variances relate to the Bangla language.

The analysis of several Bangla speech corpora, through the use of the Term Frequency - Inverse Document Frequency (TF-IDF) \cite{TF-IDF} algorithm for getting average cosine similarity scores, provides valuable insights into linguistic relationships and reveals different levels of similarity.
The Bangla and Mymensingh Speech Corpus has the most similarity, with a significant average cosine similarity score of 0.0288. Following closely after, the Bangla and Sylhet Speech Corpus had the most similarity, with a score of 0.0216. In contrast, comparisons with the Bangla and Chittagong Speech Corpus showed a lower average cosine similarity score of 0.0099. Similarly, the Bangla and Noakhali Speech Corpus shared a similarity score of 0.0139, while the Bangla and Barishal Speech Corpus had an average cosine similarity of 0.0124. These results show a decreasing linguistic similarity slope as one moves from Mymensingh and Sylhet to Noakhali, Barishal, and Chittagong.

\subsection{Translation Quality Control Process}
In our Translation Quality Control process, we implemented two essential metrics, Cohen's Kappa \cite{cohenkappa} and Fleiss' Kappa \cite{fleisskappa}, to carefully assess translation quality and inter-annotator agreement, assuring the highest level of dependability for our dataset.
Cohen's Kappa ($K_1$) was applied specifically to evaluate the translations for the Sylhet and Mymensingh regions. This metric involved the assessment of translations by Translators 1 and 2. Cohen's Kappa ($K_1$) for Sylhet and Mymensingh regions:
\begin{equation}
K_1 = 1 - \frac{1 - \kappa}{1 - \kappa_\text{max}}
\end{equation}

Where:
\begin{itemize}
\item $K_1$: Cohen's Kappa for Sylhet and Mymensingh.
\item $\kappa$: Agreement coefficient between Translators 1 and 2.
\item $\kappa_{\text{max}}$: Maximum possible agreement (usually 1).
\end{itemize}

Fleiss' Kappa ($K_2$) was employed for assessing the quality of translations in the Chittagong, Noakhali, and Barishal regions. Unlike Cohen's Kappa, Fleiss' Kappa extends the assessment to involve three Translators 1, 2, and 3. Fleiss' Kappa \(K_2\) for Chittagong, Noakhali, and Barishal regions:
\begin{equation}
\begin{split}
K_2 = \frac{1}{N(N - 1)} \Bigg[ \sum_{j=1}^{k} \left( \frac{1}{N} \sum_{i=1}^{N} n_{ij}(n_{ij} - 1) \right) \\
- \frac{1}{N(N - 1)} \sum_{j=1}^{k} \left( \sum_{i=1}^{N} n_{ij} \right)^2 \Bigg]
\end{split}
\end{equation}

Where:
\begin{itemize}
\item $K_2$: Fleiss' Kappa for Chittagong, Noakhali, Barishal.
\item $N$: Number of raters (here, 3 translators).
\item $k$: Number of rating categories.
\item $n_{ij}$: Number of raters rating the $i$th subject in the $j$th category.
\end{itemize}

In terms of translation quality, the Chittagong region demonstrated a Fleiss' Kappa rating of 0.83, while the Sylhet region showed notable agreement with a Cohen's Kappa value of 0.87. On the other hand, the Noakhali region exhibited a Fleiss' Kappa rating of 0.91, while the Mymensingh region displayed a Cohen's Kappa value of 0.92, and the Barishal region demonstrated strong unity among independent translators with a Fleiss' Kappa rating of 0.93.

\subsection{Handling Disagreements in Translation Annotation}
Disagreements among translators were inevitable due to dialectal diversity, ambiguous expressions, and stylistic choices. To address this, we employed a Double Annotation approach, where each sentence was independently translated by multiple annotators. Discrepancies were resolved through structured discussions aimed at achieving consensus based on dialectal correctness and semantic equivalence. When agreement could not be reached, decisions were made using a Weighted Voting mechanism, giving more influence to translators with higher regional fluency. In particularly complex cases, senior linguists provided final judgments. All conflicts and their resolutions were documented to support transparency and guideline refinement. These guidelines were updated iteratively to capture recurring issues and standardize future translations. This multi-step resolution framework ensured consistency, reduced subjectivity, and preserved the linguistic integrity of each regional variant in the dataset.

\subsection{Dataset Statistics}
We have carefully organized the ``Vashantor'' dataset to ensure comprehensive coverage for each region. The dataset statistics in the table below showcase the distribution of training, testing, and validation data for the five regions. Initially, we manually split the texts into 75\% for training, 15\% for testing, and 10\% for the validation set presented in Figure \ref{fig:core} and Table \ref{table4}. In Table \ref{table5}, Table \ref{table6}, and Table \ref{table7}, we provide an overview of speech corpus size, maximum text length, and minimum text length. This breakdown offers valuable insights into the variation in text lengths.

\begin{table*}
\caption{Translator Information For Bangla Regional Dialects}\label{table4}
\centering
  \begin{tabular}{p{1.8cm} p{3.5cm} p{1.5cm} p{1.2cm} p{1.8cm} p{1.4cm} } 
    \toprule
 \textbf{Region} & \textbf{Text Format} & \textbf{Train Samples} & \textbf{Test Samples} & \textbf{Validation Samples} & \textbf{Total Samples} \\
    \midrule
    {Chittagong} & Chittagong Bangla & 1875 & 375 & 250 & 2500 \\
       & Chittagong Banglish & 1875 & 375 & 250 & 2500 \\
   {Noakhali} & Noakhali Bangla & 1875 & 375 & 250 & 2500 \\ 
        & Noakhali Banglish & 1875 & 375 & 250 & 2500 \\
    {Sylhet} & Sylhet Bangla & 1875 & 375 & 250 & 2500 \\
        & Sylhet Banglish & 1875 & 375 & 250 & 2500 \\ 
{Barishal} & Barishal Bangla & 1875 & 375 & 250 & 2500 \\
        & Barishal Banglish & 1875 & 375 & 250 & 2500 \\

{Mymensingh} & Mymensingh Bangla & 1875 & 375 & 250 & 2500 \\
        & Mymensingh Banglish & 1875 & 375 & 250 & 2500 \\
    \bottomrule
  \end{tabular}
\end{table*}

\begin{table*}
   
    \caption{Dataset Length for Core Data Collection}

    \label{table5}
    \centering
    \begin{tabular}{cccc} 
        \hline
        \centering\textbf{Text Format} & \textbf{Speech Corpus Size} & \textbf{Highest Text Length} & \textbf{Lowest Text Length} \\
         & \textbf{(in words)} & \textbf{(in words)} & \textbf{(in words)} \\
        \hline
        Bangla & 72,439 & 19 & 2 \\
        Banglish & 81,514 & 19 & 2 \\
        English & 76,615 & 26 & 2 \\
       
        \hline
    \end{tabular}
\end{table*}

\begin{table*}

    \caption{Dataset Length for Different Regional Bangla Dialects}
    \label{table6}
    \centering
    \begin{tabular}{cccc}
        \hline
        \centering\textbf{Region} & \textbf{Speech Corpus Size} & \textbf{Highest Text Length} & \textbf{Lowest Text Length} \\
         & \textbf{(in words)} & \textbf{(in words)} & \textbf{(in words)} \\
        \hline
        Chittagong & 72,483 & 19 & 2 \\
        Noakhali & 72,181 & 22 & 2 \\
        Sylhet & 73,999 & 20 & 2 \\
        Barishal & 77,494 & 19 & 2 \\
        Mymensingh & 74,503 & 19 & 2 \\
        \hline
    \end{tabular}
\end{table*}

\begin{table*}
    \caption{Dataset Length for Different Regional Banglish Dialects}
    \vspace{5pt}
    \label{table7}
    \setlength\tabcolsep{5pt}
 
    \centering
    \begin{tabular}{cccc}
        \hline
        \centering\textbf{Region} & \textbf{Speech Corpus Size} & \textbf{Highest Text Length} & \textbf{Lowest Text Length} \\
         & \textbf{(in words)} & \textbf{(in words)} & \textbf{(in words)} \\
        \hline
        Chittagong & 77,599 & 19 & 2 \\
        Noakhali & 78,045 & 22 & 2 \\
        Sylhet & 82,424 & 20 & 2 \\
        Barishal & 82,587 & 19 & 2 \\
        Mymensingh & 81,751 & 19 & 2 \\
        \hline
    \end{tabular}
\end{table*}

\subsection{Benchmarking against Existing Datasets}

To establish a comprehensive regional dialect dataset, we conducted comparisons with existing datasets, including those for English to Bangla and Bangla to English translations. Notably, our ``Vashantor" dataset stands out for its distinctive incorporation of regional dialect translations. The comparative analysis in Table \ref{table8} presents a comprehensive overview of our dataset in relation to existing datasets.
\subsection{Challenges Faced}
While creating the “Vashantor” dataset, we faced various challenges that made the process more complicated. These challenges included:
\begin{itemize}
    \item Difficulty finding language experts
    \item Intra-regional language variations
    \item Diverse typing styles
    \item Spelling mistakes
\end{itemize}

People spoke in surprisingly diverse ways even within the same regions, which added another layer of complexity. A significant challenge was the translators' requirement of a thorough understanding of the languages they were translating into. They had to use caution when translating words from Bangla. For instance, when translating {\bng``sba{I}''}, (English translation: ``We") they had to make a precise decision between {\bng``ebg/Yuen''} and {\bng``ebek''}.  Translators had their unique typing styles, making consistency a challenge. An example of this is the different spellings of {\bng``khaUya''} (English translation: ``Eat") and {\bng``kha{O}ya''}, which means the same thing but are spelled differently. Dealing with these various challenges was crucial to make sure the ``Vashantor" dataset is known for its quality, accuracy, and language diversity.

\subsection{Availability and Usage} \label{Availability}
We have structured the “Vashantor” dataset in easily accessible formats, primarily available in JSON and CSV, catering to the convenience of researchers and practitioners. These formats enable easy integration into a wide range of natural language processing applications and machine learning models. Scholars and practitioners interested in utilizing the dataset can access it via our dedicated online repository, \href{https://data.mendeley.com/datasets/bj5jgk878b}{Link} providing open availability for academic and research purposes. This publicly available policy will promote the use of the dataset in language studies, dialect analysis, machine translation, and other domains. By providing straightforward access and a well-organized structure, we aim to facilitate the broadest possible usage of the “Vashantor” dataset within the research community.

\begin{table*}
\caption{Benchmarking against Existing Datasets} \label{table8}
  \begin{tabular}{p{2.0cm} p{1.5cm} p{1.5cm} p{1.6cm} p{1.4cm} p{1.2cm} p{2.5cm}} 
    \toprule
\textbf{Authors} & \textbf{Translation Directions} & \textbf{Regional Dialects to Bangla} & \textbf{Regional Dialects \newline to Banglish} & \textbf{Region \newline Detection} & \textbf{Dataset Size} & \textbf{Availability} \\
    \midrule
     Proposed work & All \newline mentioned directions & \centering Yes  & \centering Yes  & \centering Yes &  32,500 & publicly available \\
\midrule

Rafiqul \newline et al.\cite{seq} & Bangla \newline to English & \centering No & \centering No & \centering No & 9,482  & Not \newline available \\
\midrule
Shaykh \newline et al.\cite{intro5}  & English \newline to Bangla & \centering No & \centering No & \centering No & 4000 & Not \newline available \\
\midrule
Prommy \newline et al.\cite{intro6} & No & \centering No & \centering No & \centering Yes & 30 hour & Publicly available
 \\
\midrule
Kishorjit \newline et al.\cite{intro8} & Bangla \newline to Manipuri & \centering No & \centering No & \centering No & 20,687 words  & Not \newline available \\
\midrule
Niladri \newline et al.\cite{intro10} & Hindi \newline to Banglish & \centering No & \centering No & \centering No & 80000 & Not \newline available \\
\midrule

Mohammad \newline et al.\cite{MohammadBench}  & English \newline to Bangla & \centering No & \centering No & \centering No & 70,614 &  Publicly available
 \\

\midrule
Nafisa \newline et al.\cite{NafisaBenchmarks}  & Bangla \newline to English & \centering No & \centering No & \centering No & 2,660  & Not \newline available \\
    \bottomrule
  \end{tabular}
\end{table*}

\section{Dialect-to-Bangla Translation and Region Detection Models} \label{Modelslab}
\subsection{Regional Dialects to Bangla Language Translation Models}
\subsubsection{mT5}
mT5, or Massively Multilingual Pre-trained Text-to-Text Transformer \cite{mT5}, is a multilingual version of the T5 text-to-text transformer model. It is a state-of-the-art language model with a robust encoder-decoder architecture. It has been pre-trained on a vast and diverse dataset comprising 101 languages sourced from the web. mT5 comes in various model sizes, ranging from 300 million to 13 billion parameters, allowing for high-capacity and powerful language models. One of its standout features is its exceptional competence in multilingual translation tasks, making it an ideal choice for projects involving the translation of text between different languages.
\subsubsection{BanglaT5}
BanglaT5 \cite{T5} is a state-of-the-art sequence-to-sequence Transformer model designed for the Bengali language. It is based on the original Transformer architecture and has been pretrained on the extensive ``Bangla2B+" dataset, which contains 5.25 million documents gathered from a carefully selected list of web sources, totaling 27.5 GB of text data. The model architecture is the base variant of the T5 model, featuring 12 layers, 12 attention heads, a hidden size of 768, and a feed-forward size of 2048. Authors of another research  \cite{T5Nor} suggest two unique methods: aligner ensembling, which combines multiple sentence aligners to improve alignment accuracy, and batch filtering, which improves corpus quality by filtering out low-quality sentence pairings.

\subsection{Region Detection Models}
\subsubsection{mBERT}
The pretrained mBERT \cite{mBERT} model is designed for use with the top 104 languages and employs masked language modeling for self-supervised pretraining. It learns bidirectional sentence representations and sentence relationships. The publicly available model is consistent with BERT-base-cased in terms of its architectural specifications. It features 12 layers, 768 hidden units, 12 attention heads, and a total of 110 million parameters, mirroring the configuration of BERT-base-cased. mBERT can be fine-tuned on various downstream tasks and is particularly useful for tasks where the input text may be in multiple languages. It's versatile for multilingual tasks.
\subsubsection{Bangla-bert-base}
Bangla-bert-base \cite{Banglabert} is a monolingual pretrained language model that follows the BERT architecture and makes use of mask language modeling for the Bengali language. The Bengali commoncrawl corpus and the Bengali Wikipedia Dump Dataset were transformed into the BERT format, with each sentence on a separate line and an extra line to indicate document separation. The BNLP package is used to generate the model's vocabulary, which consists of 102025 tokens and is made available on GitHub and the Hugging Face model hub. The publicly available model has 12 layers, 768 hidden units, 12 attention heads, and 110 million parameters; it is consistent with the bert-base-uncased architecture. One Google Cloud GPU was used for a total of one million steps of training. An improved BERT variation titled BanglaBERT performs remarkably well through a range of Bengali NLP tasks.

\section{Evaluation Metrics} \label{EvalLab}
We explore the evaluation metrics used in this section to assess our translation models' and region detection models' performance. We categorize the evaluation into two primary components: Translation Metrics and Region Detection Metrics.
\subsection{Dialect-to-Bengali Translation Metrics}
\subsubsection{Character Error Rate}
Character Error Rate (CER) \cite{CER} is a metric used to evaluate the quality of character-level text generation. It assesses the accuracy of generated text by measuring character-level errors, including substitutions, insertions, and deletions when compared to ground truth text. The CER score is typically expressed as a percentage or fraction, with lower values indicating higher accuracy in the generated text. 

\subsubsection{Word Error Rate}
Word Error Rate (WER) \cite{WER} is a significant metric used to evaluate the accuracy and quality of generated text. WER quantifies the difference between the generated text and ground truth text in terms of words. It measures the accuracy of text output by considering word-level errors such as substitutions, insertions, deletions, and word order changes. WER is crucial for assessing the performance of translation systems. A lower WER score indicates higher accuracy, with scores closer to zero signifying that the generated translation is more faithful to the reference translation.

\begin{figure*}
\centering
\includegraphics[width=1.0\textwidth]{Regional_Speech_Big.pdf}
\caption{Overview of the Proposed Methodology for Bangla Dialect Translation and Region Detection. The diagram presents a two-stage framework: DialectBanglaT5 translates dialectal inputs into standard Bangla, while DialectBanglaBERT identifies the dialect's regional origin.}\label{fig:workflow}
\end{figure*}
\subsubsection{BLEU Score}
BLEU (Bilingual Evaluation Understudy) \cite{BLEU} is a tool for checking how good machine-generated translations are. It looks at how accurate and smooth the machine-generated text is. BLEU helps us see if the machine's translation matches human-made translations. To calculate the BLEU score, it considers factors like precision, recall, and a brevity penalty to give a complete assessment of the generated text. It works by comparing the similarity of n-grams, which are groups of n words, in the machine-generated text to the ground truth text. BLEU scores range from 0 to 1, with higher scores meaning better quality translations, especially for languages like Bengali.

\subsubsection{ROUGE Score}
We use ROUGE (Recall-Oriented Understudy for Gisting Evaluation) scores \cite{ROUGE}, which include ROUGE-1, ROUGE-2, and ROUGE-L, to evaluate machine-generated translations. These scores help us assess the quality and fluency of the translations. ROUGE-1 looks at how many single words in the machine's text match the ground truth text. ROUGE-2 checks for pairs of words (bigrams) that match, giving us a more detailed analysis of language accuracy. ROUGE-L examines the longest common sequence of words in both the machine's text and the ground truth text, providing insights into content coherence and flow.

\subsubsection{METEOR Score}

Meteor (Metric for Evaluation of Translation with Explicit ORdering) \cite{METEOR} is a powerful evaluation metric for assessing translation quality in the context of regional dialects to Bengali language translation. This metric is designed to assess the quality of translations by comparing them to human-crafted references. 

\subsection{Region Detection Metrics}
\subsubsection{Accuracy}
Accuracy \cite{Accuracy} in the context of region detection measures the proportion of correctly classified regions to the total number of regions in our dataset. It quantifies how well our model can accurately assign a text to its actual region. The score can be calculated as:
\newline
\begin{equation}
Accuracy = \frac{Number\ of\ Correctly\ Detected\ Regions}{Total\ Number\ of\ Regions}
\end{equation}

\subsubsection{Precision}
Precision \cite{PRF} is the ratio of true positives (correctly predicted instances of a specific region) to the sum of true positives and false positives (instances where the model incorrectly predicted the region). Precision is calculated as follows: 
\newline
\begin{equation}
\text{Precision} = \frac{Correctly\ Detected \ Regions} {Total \ Detected \ Regions}
\end{equation}

\subsubsection{Recall}
Recall \cite{PRF}, in the context of region detection for regional dialects, is a metric that measures the ability of a model to correctly identify and retrieve text samples belonging to a specific region from the ``Vashantor'' dataset. To put it another way, recall evaluates how well the model recognizes and categorizes text into its appropriate regional categories, ensuring that only a small number of samples are ignored or misclassified. The formula for recall is as follows:
\begin{equation}
\text{Recall} = \frac{\text{\small Correctly Detected Texts for a Region}}{\text{\small Total Texts in that Region}}
\end{equation}

\subsubsection{F1 Score}
The F1 Score \cite{PRF} for region detection is the harmonic mean of precision and recall. It is particularly useful when we want to balance the trade-off between false positives and false negatives in the context of region detection. The score can be calculated as:
\begin{equation}
\text{F1 Score} = \frac{2 \cdot \text{\small Precision} \cdot \text{\small Recall}}{\text{\small Precision} + \text{\small Recall}}
\end{equation}

\subsubsection{Logarithmic Loss}
Logarithmic Loss \cite{Logloss}, commonly known as Log Loss, is a metric used to evaluate the performance of classification models in the context of multiclass classification, where we can classify texts into one of several regions (Chittagong, Noakhali, Sylhet, Barishal, Mymensingh). It quantifies the accuracy of a model's predicted class probabilities, rewarding accurate and confident predictions while penalizing uncertain or inaccurate ones.

The Logarithmic Loss (Log Loss) for region detection is calculated as:
\begin{equation}
\text{Log Loss} = -\frac{1}{N} \sum \left[ y \log(p) + (1 - y) \log(1 - p) \right]
\end{equation}
\begin{align*}
\text{Where:} \\
N &: \text{Total number of texts.} \\
y &: \text{True label (1 if text is in region, 0 otherwise).} \\
p &: \text{Predicted probability for the region.}
\end{align*}




\section{Proposed Methodology} \label{Methodologylab}

This section outlines the proposed methodology for addressing the dual tasks of handling regional Bangla dialects: translating regional dialects into standard Bangla and identifying the dialect’s region of origin. We introduce two novel models: (1) DialectBanglaT5, designed for translating dialectal texts to standard Bangla, and (2) DialectBanglaBERT, developed for region detection. The methodology leverages the Vashantor dataset, which includes dialectal Bangla sentences from five regions (Chittagong, Sylhet, Noakhali, Barishal, and Mymensingh), such as {\bng Aen ik cn/na Aa{I} EeDt/tun cil ja{I}?''} (Chittagong), {\bng Aaen/n ik can Aa{I} EeDt/tun cil Ja{I}?''} (Noakhali), {\bng Aaphen ca{I}n in Aaim {I}n taik Ja{I} ig?''} (Sylhet), {\bng Aamen ik can mu{I} E{I}Haen egaen c{I}lLa Ja{I}?''} (Barishal), and {\bng Aaphen ikta can Aaim E{I}Han tha{I}k/ka c{I}lLa Ja{I}?''} (Mymensingh), all translating to Do you want me to leave here?'' in English. The workflow, detailed in Figure \ref{fig:workflow} and the subsequent sections, is evaluated using multiple performance metrics, including BLEU, METEOR, ROUGE, CER, and WER for translation, and Accuracy, Precision, Recall, F1-score, and Log Loss for region detection, to assess the effectiveness of both models against baseline approaches.\\ 


\subsection{Regional Dialect-to-Standard Bangla Translation}
This section describes the proposed methodology for translating different regional dialects to their corresponding standard Bengali language and region detection task, which is broken down into nine primary phases.\\

\textbf{\customunderline{Phase 1) Input Text:}}
We begin by extracting dialectal Bangla texts from the Vashantor dataset. Each text sample is normalized through standard preprocessing techniques including punctuation normalization, Unicode standardization, stop word removal, and whitespace cleaning.

\textbf{\customunderline{Phase 2) Obtain Human Translation:}}
During this phase, we collect human-generated translations from regional dialects into standard Bangla. These human translations act as the foundation for accurate translation by serving as the ground truth. While regional dialects vary greatly across Chittagong, Noakhali, Sylhet, Barishal, and Mymensingh, the human-generated Bangla translations are stable throughout all of them. These translations serve as a universal bridge, ensuring that the final output is in standard Bangla regardless of the original text's regional dialect. This phase ensures that the translation process follows a common, recognized Bangla language to provide clear and precise communication.

\textbf{\customunderline{Phase 3) Regional Dialect Translation Model Architec-\\}} \textbf{\customunderline{ture (DialectBanglaT5):}} DialectBanglaT5 is a customized extension of the BanglaT5 encoder-decoder architecture, designed specifically to address the complexity of regional Bangla dialects. To enhance the model’s alignment and fluency in translating informal or dialectal structures, we introduce two major architectural enhancements in the decoder: an additional cross-attention layer and a learnable gating mechanism.

After the standard decoding process, we append an extra decoder block that includes multi-head cross-attention over the encoder outputs. This secondary attention step enables the decoder to re-attend to source-side information, allowing it to revise and refine its understanding of contextually rich or ambiguous dialectal tokens. To effectively integrate this new representation, we use a gating mechanism that softly fuses the original decoder output with the enhanced context from the added attention layer.

Formally, let \( Z \in \mathbb{R}^{t \times d} \) be the decoder hidden states, and \( E \in \mathbb{R}^{s \times d} \) the encoder outputs, where \( t \) and \( s \) are the target and source sequence lengths, respectively, and \( d \) is the hidden size. We compute:
\begin{align*}
Z' &= \text{ExtraDecoderCrossAttention}(Z, E)  \\
G &= \sigma(W_g \cdot Z + b_g) \in \mathbb{R}^{t \times d} \\
Z_{\text{fused}} &= G \odot Z' + (1 - G) \odot Z \\
\text{logits} &= \text{LMHead}(Z_{\text{fused}})
\end{align*}


This enhanced decoder architecture enables DialectBanglaT5 to better align rare dialectal inputs with standard Bangla targets, improving translation accuracy and fluency. We train one instance of DialectBanglaT5 for each dialectal region (Chittagong, Sylhet, Noakhali, Barishal, and Mymensingh) to ensure region-specific specialization.

\textbf{\customunderline{Phase 4) Comparative Translation Baselines:}}
 In addition to DialectBanglaT5, we train BanglaT5 and mT5-base using the same dataset and evaluation setup. BanglaT5 serves as a strong monolingual baseline, while mT5-base provides insights from a multilingual model. Our proposed DialectBanglaT5, equipped with an additional cross-attention layer and gated output fusion, achieves consistently better translation performance across all dialects when compared to both BanglaT5 and mT5-base.

\textbf{\customunderline{Phase 5) Hyperparameter Tuning:}}
At this phase, we focus on improving our models' performance by fine-tuning their hyperparameters. Hyperparameters are external configuration variables that control how our translation and models perform. We want to increase the accuracy, efficiency, and overall quality of our models by improving these hyperparameters. In Section \ref{Hyperparameter}, we present the detailed hyperparameter tuning procedures for translation models.

\textbf{\customunderline{Phase 6) Generation of Translation Options:}}
Using the two different models, we generate two alternative translations for each input text. This offers us ten possible translations for the five regional dialects. This method allows us to analyze various translation options and select the one that most closely represents the standard Bangla language.

\textbf{\customunderline{Phase 7) Post-processing Enhancement:}} 
We intend to improve our translations and region detection in this phase. We accomplish this by carefully enhancing the translated text using specialized methods. These methods examine grammar, punctuation, and the text's overall soundness. These complex methods entail looking at the entire text to ensure consistency and that the translation sounds authentic. In addition, we edit any grammar errors and change the style and tone to match the situation at hand. 

\textbf{\customunderline{Phase 8) Translation Quality Assessment:}} We apply five types of metrics to determine the quality of our translations from regional dialects to Bangla. These metrics assist us in measuring various aspects of translation accuracy and fluency. These metrics include: CER, WER, BLEU, METEOR, ROUGE(ROUGE-1, ROUGE-2, and ROUGE-L). In Section \ref{Experiments}, we dig into an in-depth analysis of the scores obtained from these metrics. This analysis compares the performance of individual translation models and gives qualitative insights into translation quality.

\begin{algorithm}
\caption{DialectBanglaT5 Translation}
\begin{algorithmic}[1]
\Require Dialectal sentence $X$, Trained DialectBanglaT5 parameters $\theta$, Tokenizer $T$
\Ensure Translated sentence $\hat{Y}$ in standard Bangla
\State Preprocess $X$: normalize punctuation, remove stopwords, convert to Unicode
\State Tokenize input: $X_{\text{ids}} \gets T.\text{tokenize}(X)$
\State Encode input: $E \gets \text{Encoder}(X_{\text{ids}})$
\State Initialize decoder input: $Y_0 \gets \text{start token}$
\For{each decoding step $t = 1$ to $\text{max\_length}$}
    \State $Z_t \gets \text{Decoder}(Y_{t-1}, \text{encoder\_outputs} = E)$
    \State $Z'_t \gets \text{ExtraDecoderCrossAttention}(Z_t, E)$
    \State $G_t \gets \sigma(W_g \cdot Z_t + b_g)$
    \State $Z_{\text{fused}} \gets G_t \odot Z'_t + (1 - G_t) \odot Z_t$
    \State $\text{logits} \gets \text{LMHead}(Z_{\text{fused}})$
    \State $Y_t \gets \arg\max(\text{logits})$
    \If{$Y_t$ is end-of-sequence}
        \State \textbf{break}
    \EndIf
\EndFor
\State Detokenize output: $\hat{Y} \gets T.\text{detokenize}(\{Y_1, Y_2, \dots, Y_t\})$
\State \Return $\hat{Y}$
\end{algorithmic}
\end{algorithm}

\begin{algorithm}
\caption{DialectBanglaBERT Region Detection}
\begin{algorithmic}[1]
\Require Dialectal sentence $X$, Trained DialectBanglaBERT parameters $\phi$, Tokenizer $T$
\Ensure Predicted region label $\hat{R} \in \{\text{Chittagong}, \text{Sylhet}, \text{Noakhali}, \text{Barishal}, \text{Mymensingh}\}$
\State Preprocess $X$: clean text, normalize, remove noise
\State Tokenize input: $X_{\text{ids}} \gets T.\text{tokenize}(X)$
\State Pass through Bangla-BERT: $H \gets \text{BanglaBERT}(X_{\text{ids}})$ \Comment{$H \in \mathbb{R}^{n \times d}$}
\State Refine representation: $H' \gets \text{MultiHeadSelfAttention}(H)$
\State Aggregate sequence: $h_{\text{cls}} \gets \text{MeanPooling}(H')$
\State Compute region logits: $\text{logits} \gets \text{Classifier}(h_{\text{cls}})$
\State Apply softmax: $P \gets \text{softmax}(\text{logits})$
\State $\hat{R} \gets \arg\max(P)$
\State \Return $\hat{R}$
\end{algorithmic}
\end{algorithm}
\subsection{Region Detection from Regional Dialect}

\textbf{\customunderline{Phase 1) Dataset Construction:}} We use region-annotated dialectal text samples from the Vashantor dataset for training and evaluation. Each sample is preprocessed as in the translation pipeline and assigned a label based on its dialectal origin. 

\textbf{\customunderline{Phase 2)  Region Detection Model Architecture:}}

DialectBanglaBERT builds upon the Bangla-BERT backbone by integrating an additional self-attention layer above the final transformer block. This extra attention mechanism is designed to re-contextualize the final hidden states by refining token-level interactions before classification. Instead of directly using the \texttt{[CLS]} token output, the model applies an additional multi-head self-attention layer over the final layer’s token embeddings, followed by average pooling and a fully connected classification head.

Formally, let $H \in \mathbb{R}^{n \times d}$ be the output of the final Bangla-BERT layer for a sequence of $n$ tokens with embedding size $d$. The model computes:
\begin{align*}
H' &= \text{MultiHeadSelfAttention}(H) \\
h_{\text{cls}} &= \text{MeanPooling}(H') \\
\text{logits} &= \text{FC}(h_{\text{cls}})
\end{align*}

This architectural enhancement allows the model to capture nuanced inter-token dependencies specific to dialectal expressions, which are often subtle and context-sensitive. As a result, the model demonstrates improved performance over standard transformer-based classifiers in distinguishing regional Bangla varieties. 

\textbf{\customunderline{Phase 3)  Baselines comparison for Region Detection:}} 
For comparative analysis, we fine-tune mBERT and BanglaBERT on the same region-labeled data. While both models can capture broad linguistic features, they are less effective in distinguishing subtle dialectal variations. Our proposed DialectBanglaBERT, enhanced with an additional attention layer, achieves consistently better performance across all evaluation metrics. 

 \begin{table*}
\caption{Hyperparameter Tuning for Bangla Regional Dialects to Bangla Translation}\label{table12}
\centering
\small
  \begin{tabular}{l p{2.0cm} p{1.5cm} p{1.4cm} p{1.2cm} p{1.4cm} p{1.2cm}} 
    \toprule
 \textbf{Region} & \textbf{Models} & \textbf{Learning Rate} & \textbf{Batch Size} & \textbf{Number of Epochs} & \textbf{Optimizer} & \textbf{Sequence Length} \\
    \midrule
    {Chittagong} & mT5 & 0.001 & 16 & 50 & AdamW & 128 \\
       & BanglaT5 & 0.001 & 16 & 53 & AdamW & 128 \\
        &  DialectBanglaT5 & 0.001 & 16 & 45 & AdamW & 128 \\
     \midrule
   {Noakhali} & mT5  & 0.001 & 16 & 45 & AdamW & 128 \\
        & BanglaT5 & 0.001 & 16 & 40 & AdamW & 128 \\
        & DialectBanglaT5 & 0.001 & 16 & 40 & AdamW & 128 \\
 \midrule
    {Sylhet} & mT5 & 0.001 & 16 & 43 & AdamW & 128 \\
        & BanglaT5 & 0.001 & 16 & 45 & AdamW & 128 \\
        & DialectBanglaT5 & 0.001 & 16 & 40 & AdamW & 128 \\
         \midrule
{Barishal} & mT5 & 0.001 & 16 & 35 & AdamW & 128 \\
        & BanglaT5 & 0.001 & 16 & 35 & AdamW & 128 \\
        & DialectBanglaT5 & 0.001 & 16 & 32 & AdamW & 128 \\
\midrule
{Mymensingh} & mT5 & 0.001 & 16 &  30 & AdamW & 128 \\
        & BanglaT5 & 0.001 & 16 & 28 & AdamW & 128 \\
        & DialectBanglaT5 & 0.001 & 16 & 25 & AdamW & 128 \\
    \bottomrule
  \end{tabular}
\end{table*}

\begin{table*}
\caption{Hyperparameter Tuning for Region Detection}\label{table13}
\centering
\small 
  \begin{tabular}{l l p{1.2cm} p{1.2cm} p{1.2cm} p{1.4cm} p{1.2cm}} 
    \toprule
 \textbf{Region} & \textbf{Models} & \textbf{Learning Rate} & \textbf{Batch Size} & \textbf{Number of Epochs} & \textbf{Optimizer} & \textbf{Sequence Length} \\
    \midrule
    {Chittagong} & mBERT & 0.00002 & 16 & 10 & AdamW & 128 \\
       & Bangla-bert-base  & 0.00002 & 16 & 10 & AdamW & 128 \\
       & DialectBanglaBERT  & 0.00002 & 16 & 10 & AdamW & 128 \\
       
     \midrule
   {Noakhali} & mBERT  & 0.00002 & 16 & 10 & AdamW & 128 \\
        & Bangla-bert-base  & 0.00002 & 16 & 10 & AdamW & 128 \\
        & DialectBanglaBERT  & 0.00002 & 16 & 10 & AdamW & 128 \\
 \midrule
    {Sylhet} & mBERT & 0.00002 & 16 & 10 & AdamW & 128 \\
        & Bangla-bert-base  & 0.00002 & 16 & 10 & AdamW & 128 \\
        & DialectBanglaBERT  & 0.00002 & 16 & 10 & AdamW & 128 \\
         \midrule
{Barishal} & mBERT & 0.00002 & 16 & 10 & AdamW & 128 \\
        & Bangla-bert-base  & 0.00002 & 16 & 10 & AdamW & 128 \\
        & DialectBanglaBERT  & 0.00002 & 16 & 10 & AdamW & 128 \\
\midrule
{Mymensingh} & mBERT & 0.00002 & 16 &  10 & AdamW & 128 \\
        & Bangla-bert-base  & 0.00002 & 16 & 10 & AdamW & 128 \\
        & DialectBanglaBERT  & 0.00002 & 16 & 10 & AdamW & 128 \\
    \bottomrule
  \end{tabular}
\end{table*}
\textbf{\customunderline{Phase 4) Evaluation of Region Detection:}} During this phase, we want to ensure that our models can correctly identify the region from where the input text originates. We utilize many metrics to assess how effectively it operates. Accuracy, Precision, Recall, F1 Score, and Log Loss are examples of these metrics. In Section \ref{Experiments}, We perform a complete metric score analysis to evaluate and compare the performance of different region detection models, providing significant insights into their accuracy and effectiveness.

\begin{table*}
\caption{CER, WER, BLEU, METEOR scores of all the Bangla regional dialect translation models}
\label{table9}
\centering
  \begin{tabular}{llcccc}
    \toprule
\textbf{Region} & \textbf{Models} & \textbf{CER} & \textbf{WER} & \textbf{BLEU} & \textbf{METEOR}  \\
    \midrule
    {Chittagong} & mT5 & 0.2308 & 0.3959 & 36.75 & {0.6008}\\
 & BanglaT5 & {0.2040} & 0.3385 & 44.03 & {0.6589} \\
 & \textbf{DialectBanglaT5} & {0.2009} & 0.3225 & \textbf{46.91} & {0.6750} \\
    \midrule
   {Noakhali} & mT5 & 0.2035 & 0.3870 & {37.43} & {0.6073}\\
 & BanglaT5 & {0.1863} & {0.3214} & 47.38  & {0.6802} \\
 & \textbf{DialectBanglaT5} & 0.1821 & 0.3086 & \textbf{49.85}  & 0.6985 \\
\midrule
    {Sylhet} & mT5 & {0.1472} & {0.2695} & 51.32 & {0.7089}\\
 & BanglaT5 & 0.1715 & 0.2802 & 51.08  & 0.7073\\
 & \textbf{DialectBanglaT5} & 0.1420 & 0.2558 & \textbf{53.87} & 0.7284 \\
\midrule
{Barishal} &  mT5 & {0.1480} & {0.2644} & {48.56}  & {0.7175}\\
 & BanglaT5 & 0.1497 & 0.2459 & 53.50 & 0.7334\\
 & \textbf{DialectBanglaT5} & 0.1425 & 0.2331 & \textbf{55.94} & 0.7528 \\
\midrule
{Mymensingh} & mT5 & 0.0796 & 0.1674 & 64.74 & {0.8201}\\
 & BanglaT5 & 0.0823 & 0.1548 & 69.06  & 0.8312\\
 & \textbf{DialectBanglaT5} & 0.0791 & 0.1470 & \textbf{71.93} & 0.8503 \\
    \bottomrule
  \end{tabular}
\end{table*}
\begin{figure*}
\centering
\includegraphics[width=1.0\textwidth]{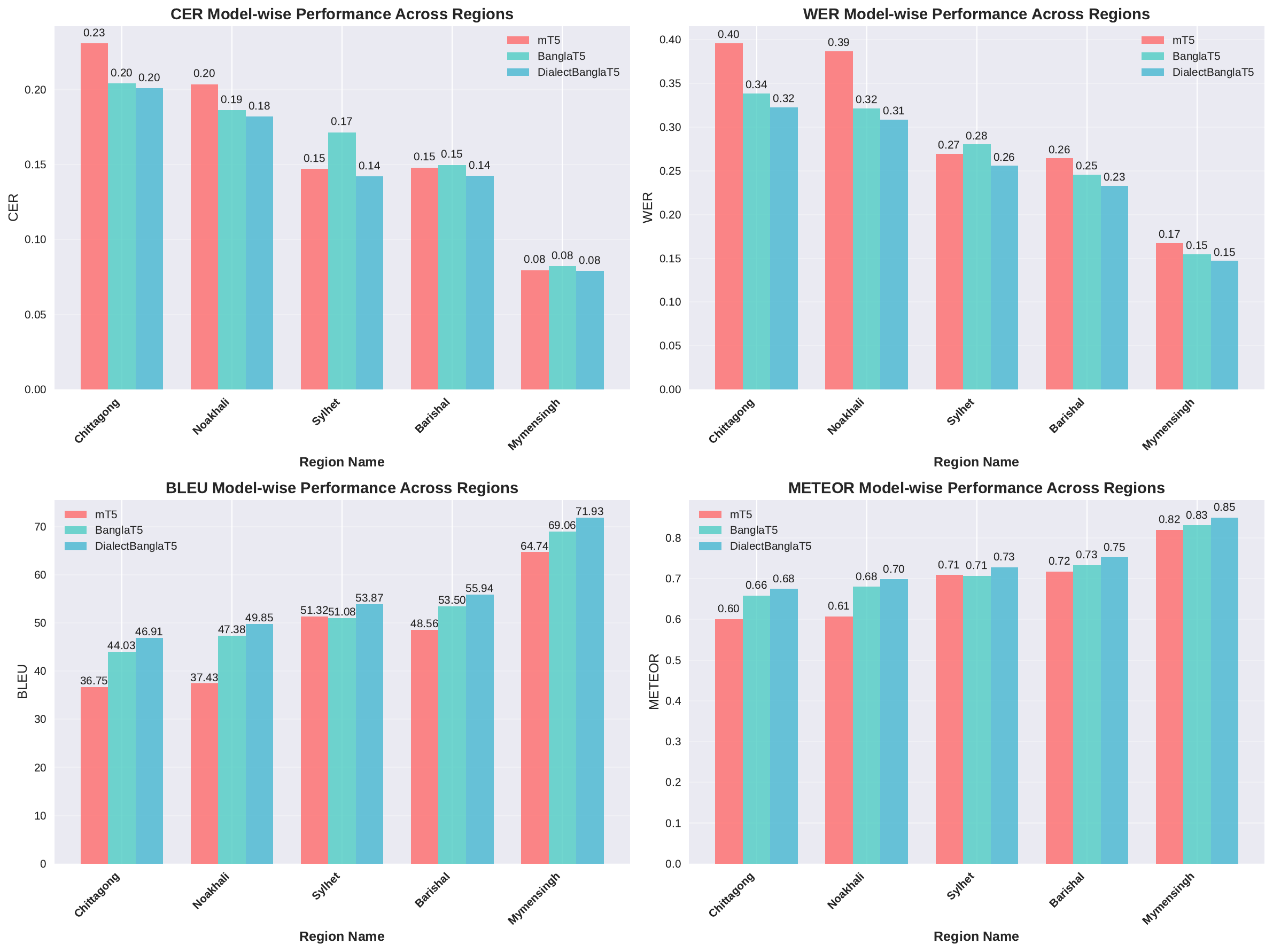}
\caption{Model-wise performance comparison across five Bangla dialect regions (Chittagong, Noakhali, Sylhet, Barishal, and Mymensingh) using four evaluation metrics: CER, WER, BLEU, and METEOR. }\label{model_performance_comparison}
\end{figure*}
\begin{figure*}
\centering
\includegraphics[width=1.0\textwidth]{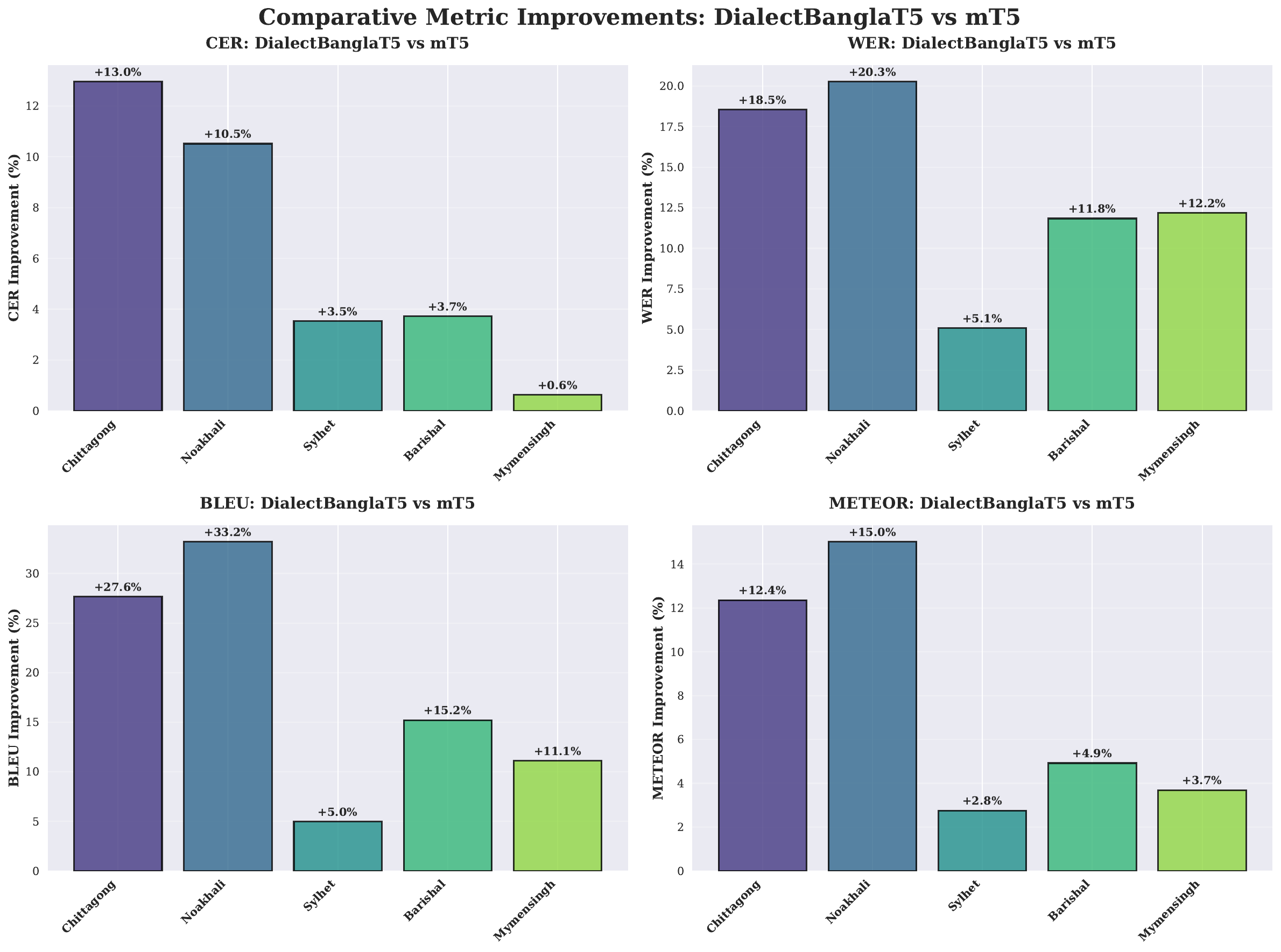}
\caption{Metric-wise comparative improvements of DialectBanglaT5 over mT5 across five Bangla regional dialects. DialectBanglaT5 shows consistent gains in CER, WER, BLEU, and METEOR.}\label{DialectBanglaT5vsmt5}
\end{figure*}
\begin{figure*}
\centering
\includegraphics[width=1.0\textwidth]{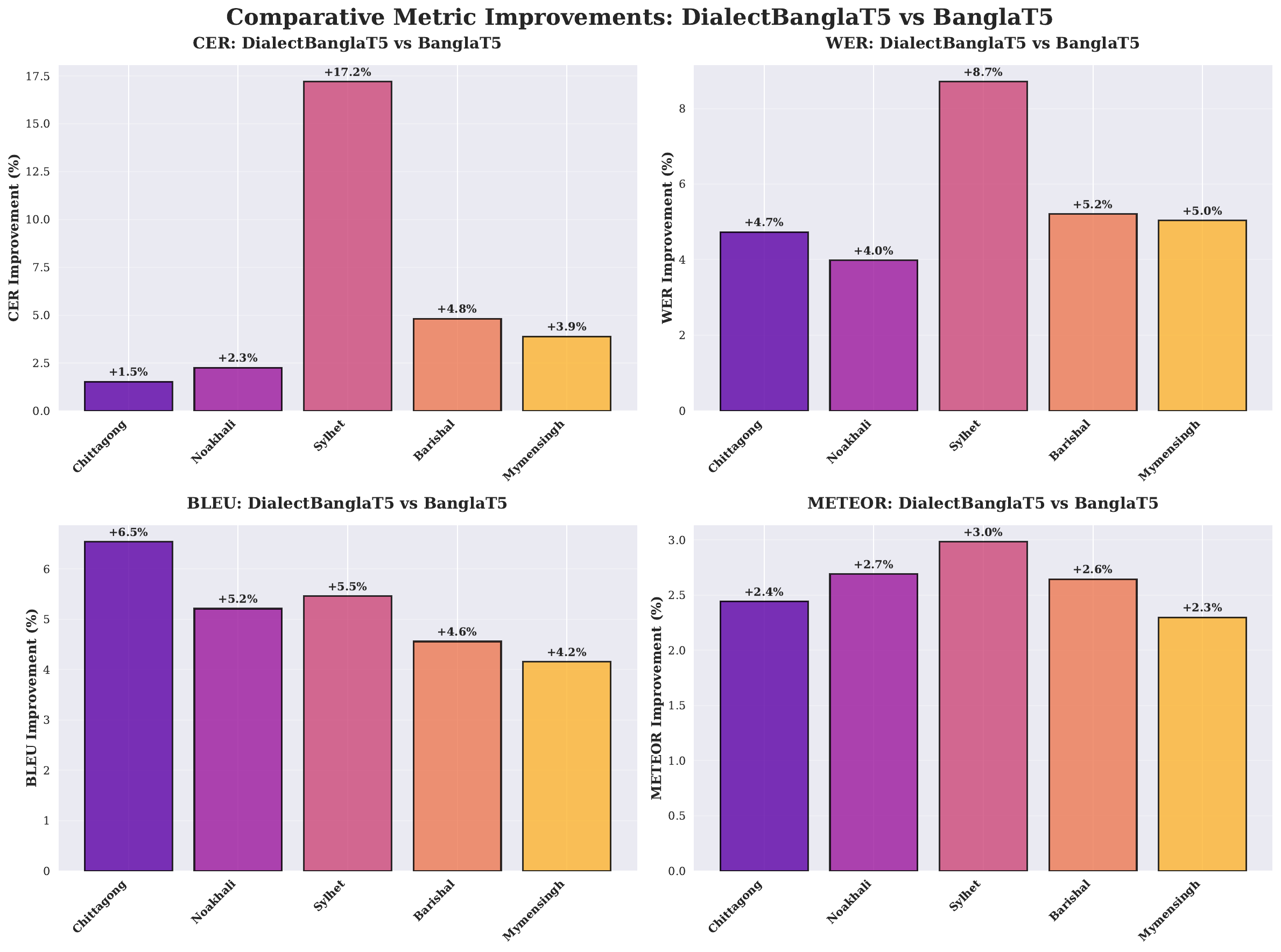}
\caption{Metric-wise comparative improvements of DialectBanglaT5 over BanglaT5 across five Bangla regional dialects. DialectBanglaT5 shows consistent gains in CER, WER, BLEU, and METEOR.}\label{DialectBanglaT5vsBanglat5}
\end{figure*}
\begin{figure}
\centering
\includegraphics[width=1.0\linewidth]{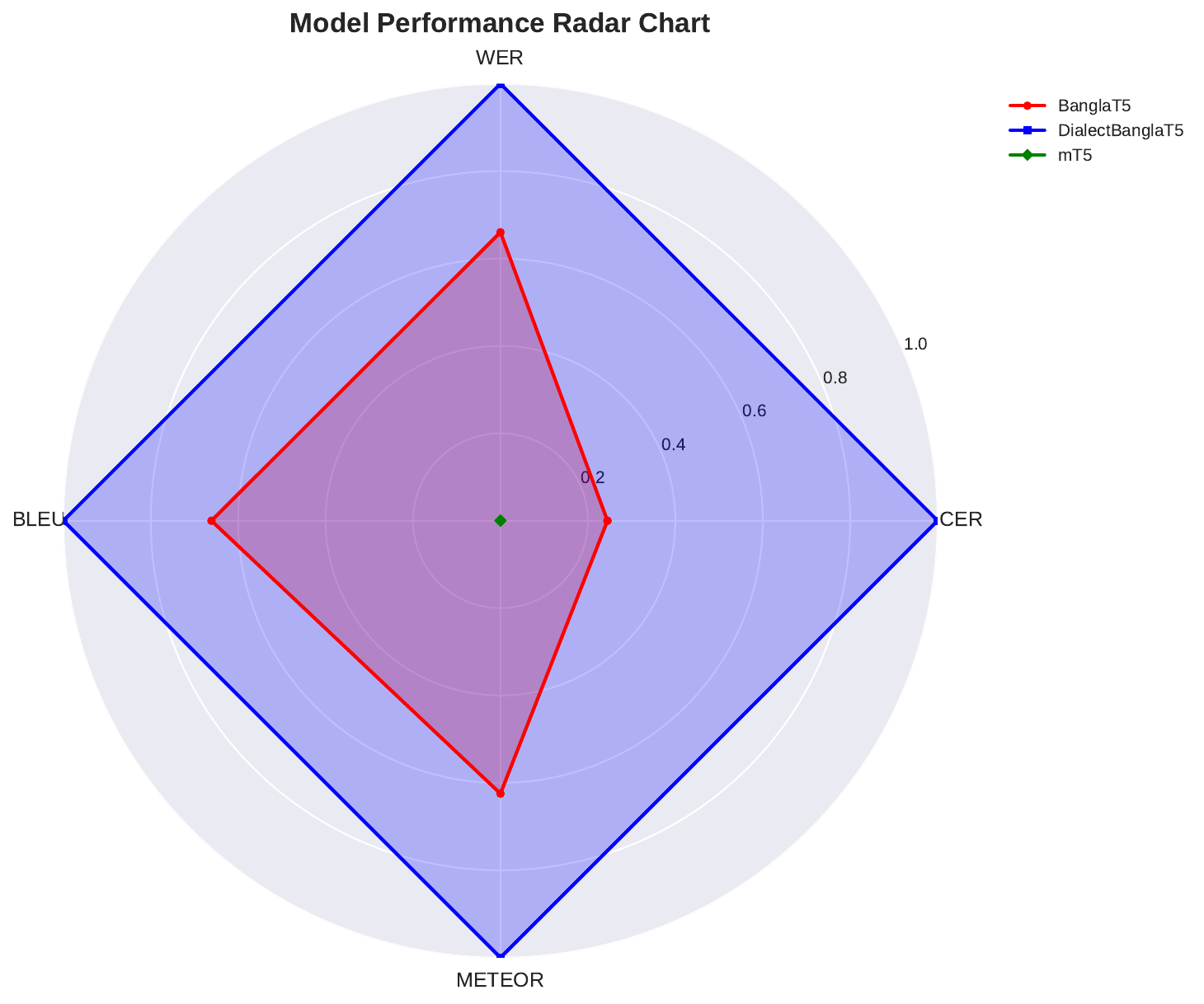}
\caption{Radar Chart Comparing Translation Model Performance on Dialectal Bangla Benchmarks
This radar chart presents the performance comparison of three models: BanglaT5, DialectBanglaT5, and mT5, evaluated using four standard metrics namely WER, CER, BLEU, and METEOR.}\label{Radar_Capped}
\end{figure}
\section{Experimental Results and Analysis} \label{Resultslab}
\subsection{Experimental Setup}
The experiments were conducted on two different setups. The first setup used Google Colaboratory, with Python 3.10.12, PyTorch 2.0.1, a Tesla T4 GPU (15 GB), 12.5 GB of RAM, and 64 GB of disk space. The second setup used the Jupyter Notebook environment, with Python 3.10.12, PyTorch 2.0.1, an NVIDIA GeForce RTX 3050 GPU (8 GB), 16 GB of RAM, and a 512 GB NVMe SSD.

\subsection{Hyperparameter Settings} \label{Hyperparameter}
The Table \ref{table12} shows the hyperparameters used to train a regional dialects translation model for five different regions in Bangladesh: Chittagong, Noakhali, Sylhet, Barishal, and Mymensingh. In the hyperparameter tuning for Bangla regional dialects to Bangla translation using two models, mT5 and BanglaT5. Key hyperparameters include a learning rate of 0.001, a fixed batch size of 16, and varying numbers of epochs for each region and model, such as the highest number of epochs is observed in the Chittagong region for the BanglaT5 model, reaching 53, while the lowest is found in the Mymensingh region for the same model, with 28 epochs. The optimization algorithm employed is AdamW. Additionally, a sequence length of 128 is set, a critical parameter for tasks dealing with sequential data like natural language processing. Moving on to Table \ref{table13}, the hyperparameter tuning results for region detection are presented, focusing on two pre-trained BERT models: mBERT and Bangla-Bert-Base. For all regions, both models are trained with consistent hyperparameter values, including a learning rate of 0.00002, a batch size of 16, 10 epochs using the AdamW optimizer, and a sequence length of 128.


\subsection{Experiments}\label{Experiments}


\begin{table*}
\caption{ROUGE scores of all the Bangla regional dialect translation models}\label{table10}
\centering
\begin{tabular}{llccc ccc ccc}
    \toprule
    \multirow{2}{*}{\textbf{Region}} & \textbf{Model} & 
    \multicolumn{3}{c}{\textbf{ROUGE-1}} & \multicolumn{3}{c}{\textbf{ROUGE-2}} & \multicolumn{3}{c}{\textbf{ROUGE-L}} \\ 
    \cmidrule(lr){3-5} \cmidrule(lr){6-8} \cmidrule(lr){9-11}
     & & \textbf{R} & \textbf{P} & \textbf{F} 
     & \textbf{R} & \textbf{P} & \textbf{F}  
     & \textbf{R} & \textbf{P} & \textbf{F}  \\
    \midrule
    {Chittagong} & mT5 & 0.6563 & 0.6820 & 0.6659 & 0.4662 & 0.4854 & 0.4733 & 0.6526 & 0.6784 & 0.6623 \\
    & BanglaT5 & 0.7082 & 0.7321 & 0.7172 & 0.5217 & 0.5413 & 0.5290 & 0.7032 & 0.7272 & 0.7123 \\
    & \textbf{DialectBanglaT5} & 0.7200 & 0.7468 & 0.7325 & 0.5335 & 0.5520 & 0.5385 & 0.7160 & 0.7417 & 0.7260 \\
    \midrule
    {Noakhali} & mT5 & 0.6670 & 0.6753 & 0.6642 & 0.4765 & 0.4745 & 0.4712 & 0.6615 & 0.6723 & 0.6642 \\
    & BanglaT5 & 0.7282 & 0.7312 & 0.7245 & 0.5517 & 0.5632 & 0.5590 & 0.7221 & 0.7321 & 0.7232 \\
    & \textbf{DialectBanglaT5} & 0.7400 & 0.7458 & 0.7385 & 0.5635 & 0.5745 & 0.5685 & 0.7342 & 0.7439 & 0.7365 \\
    \midrule
    {Sylhet} & mT5 & 0.7487 & 0.7721 & 0.7584 & 0.5851 & 0.6028 & 0.5923 & 0.7472 & 0.7703 & 0.7568 \\
    & BanglaT5 & 0.7493 & 0.7721 & 0.7578 & 0.5881 & 0.6054 & 0.5944 & 0.7477 & 0.7705 & 0.7562 \\
    & \textbf{DialectBanglaT5} & 0.7620 & 0.7845 & 0.7715 & 0.5995 & 0.6165 & 0.6055 & 0.7604 & 0.7829 & 0.7698 \\
    \midrule
    {Barishal} & mT5 & 0.7545 & 0.7635 & 0.7628 & 0.5877 & 0.5968 & 0.5885 & 0.7524 & 0.7623 & 0.7585 \\
    & BanglaT5 & 0.7735 & 0.7759 & 0.7729 & 0.6084 & 0.6107 & 0.6082 & 0.7732 & 0.7755 & 0.7726 \\
    & \textbf{DialectBanglaT5} & 0.7890 & 0.7915 & 0.7886 & 0.6205 & 0.6230 & 0.6203 & 0.7888 & 0.7912 & 0.7885 \\
    \midrule
    {Mymensingh} & mT5 & 0.8418 & 0.8458 & 0.8431 & 0.7176 & 0.7214 & 0.7189 & 0.8418 & 0.8458 & 0.8431 \\
    & BanglaT5 & 0.8407 & 0.8355 & 0.8362 & 0.7128 & 0.7088 & 0.7091 & 0.8407 & 0.8355 & 0.8362 \\
    & \textbf{DialectBanglaT5} & 0.8570 & 0.8508 & 0.8525 & 0.7285 & 0.7235 & 0.7248 & 0.8570 & 0.8508 & 0.8525 \\
    \bottomrule
\end{tabular}
\end{table*}
\begin{figure*}
    \centering
    \begin{subfigure}{0.32\textwidth}
        \includegraphics[width=\linewidth]{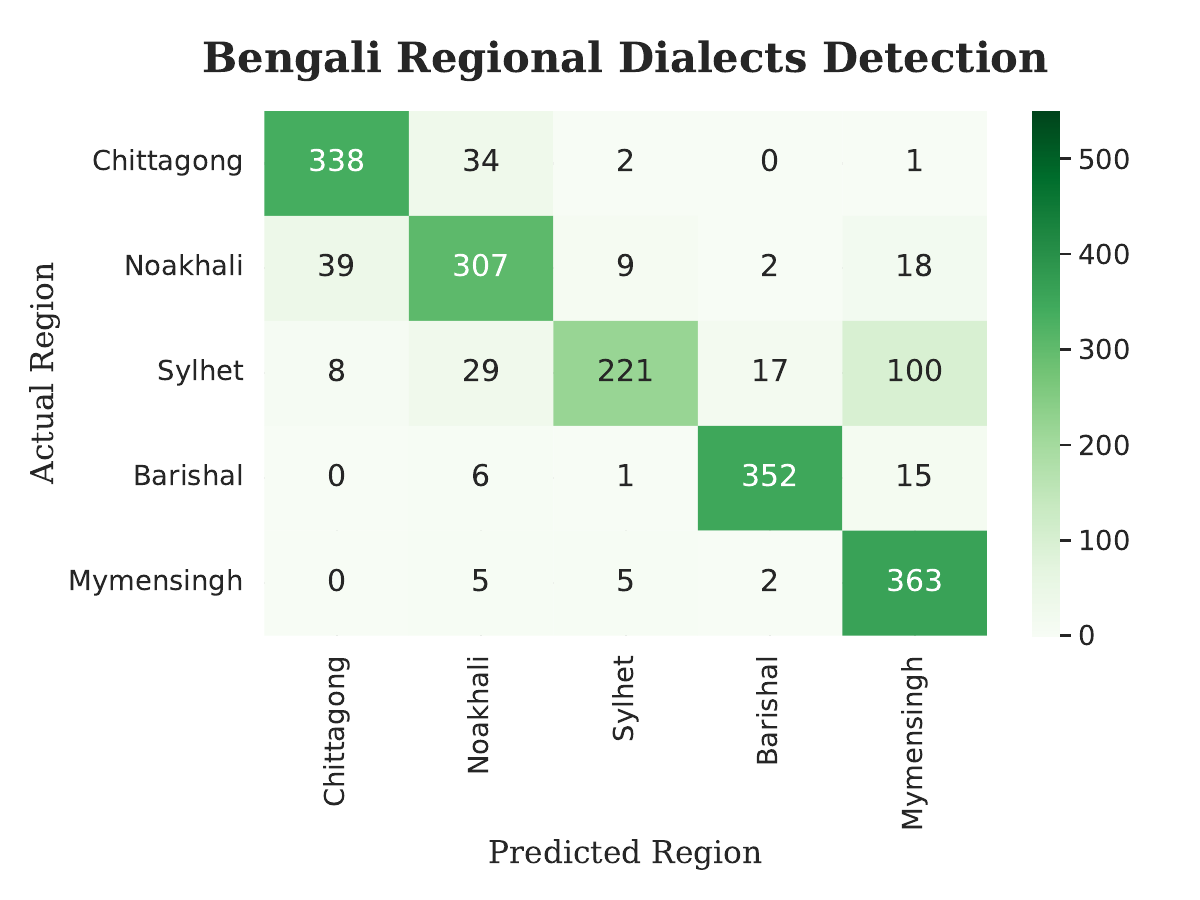}
        \caption{mBERT}
        \label{figconf1}
    \end{subfigure}
    \hfill
    \begin{subfigure}{0.32\textwidth}
        \includegraphics[width=\linewidth]{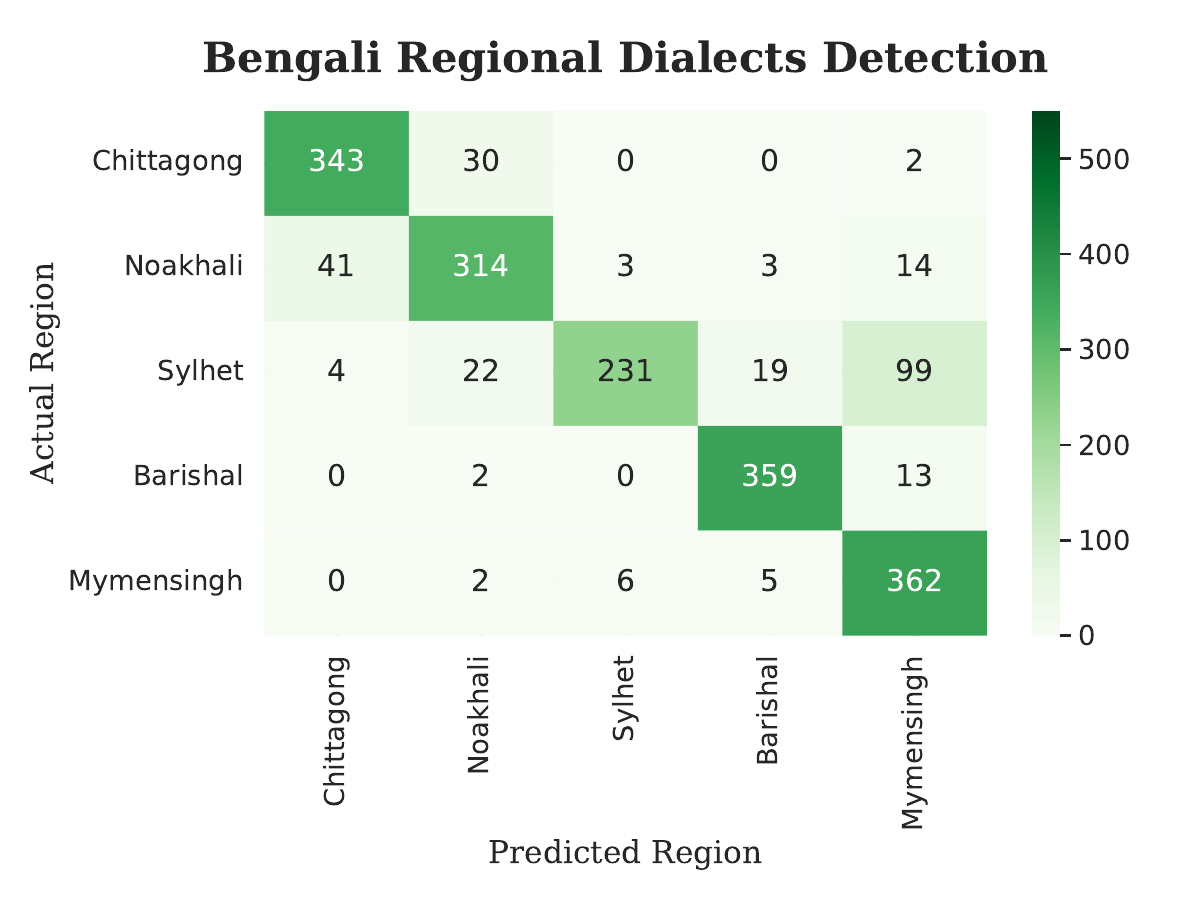}
        \caption{Bangla-bert-base} 
        \label{figconf2}
    \end{subfigure}
    \hfill
    \begin{subfigure}{0.32\textwidth}
        \includegraphics[width=\linewidth]{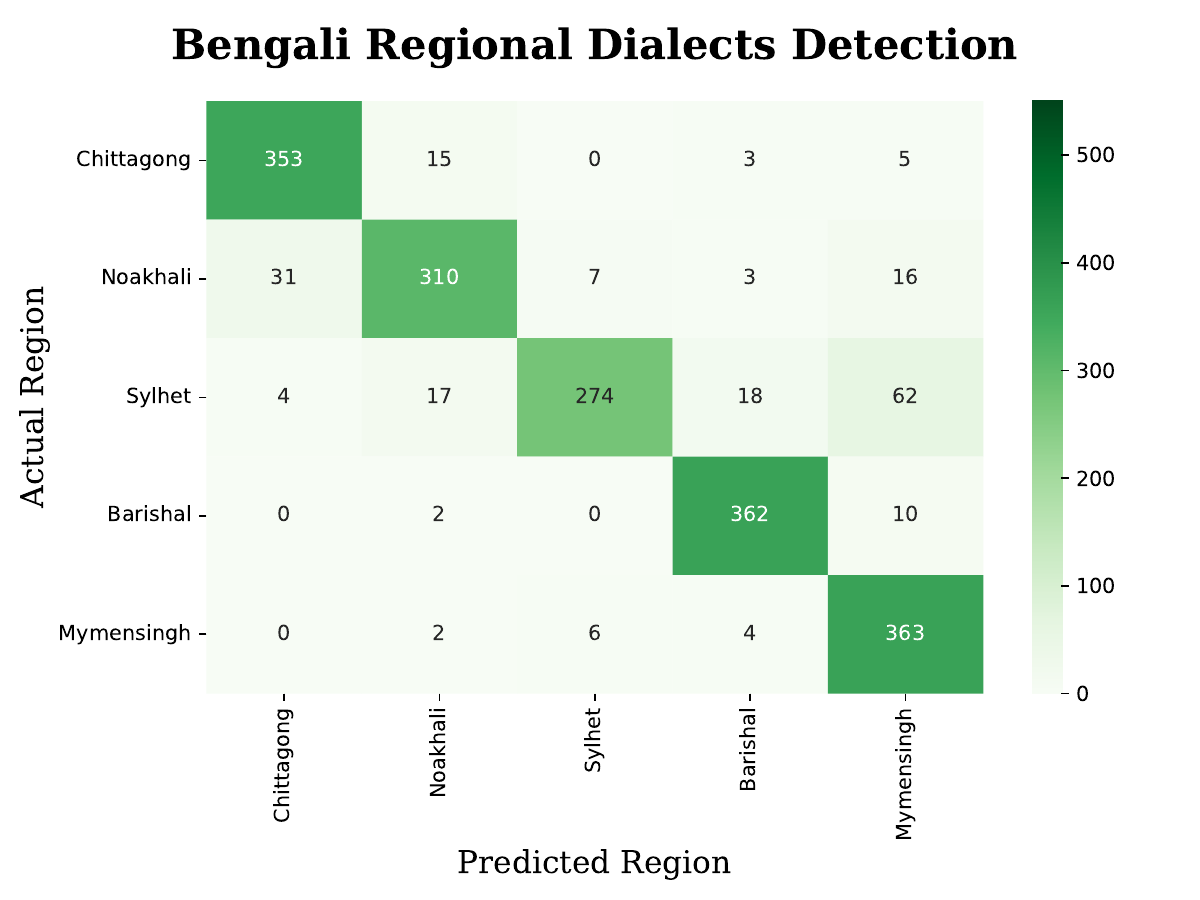}
        \caption{DialectBanglaBERT}
        \label{figconf3}
    \end{subfigure}

    \caption{Confusion matrix heatmaps for regional dialect classification using three models: (a) mBERT, (b) BanglaBERT-base, and (c) DialectBanglaBERT.}\label{fig:fig1andfig2}

\end{figure*}

\begin{figure*}
    \centering
    \begin{subfigure}{0.32\textwidth}
        \includegraphics[width=\linewidth]{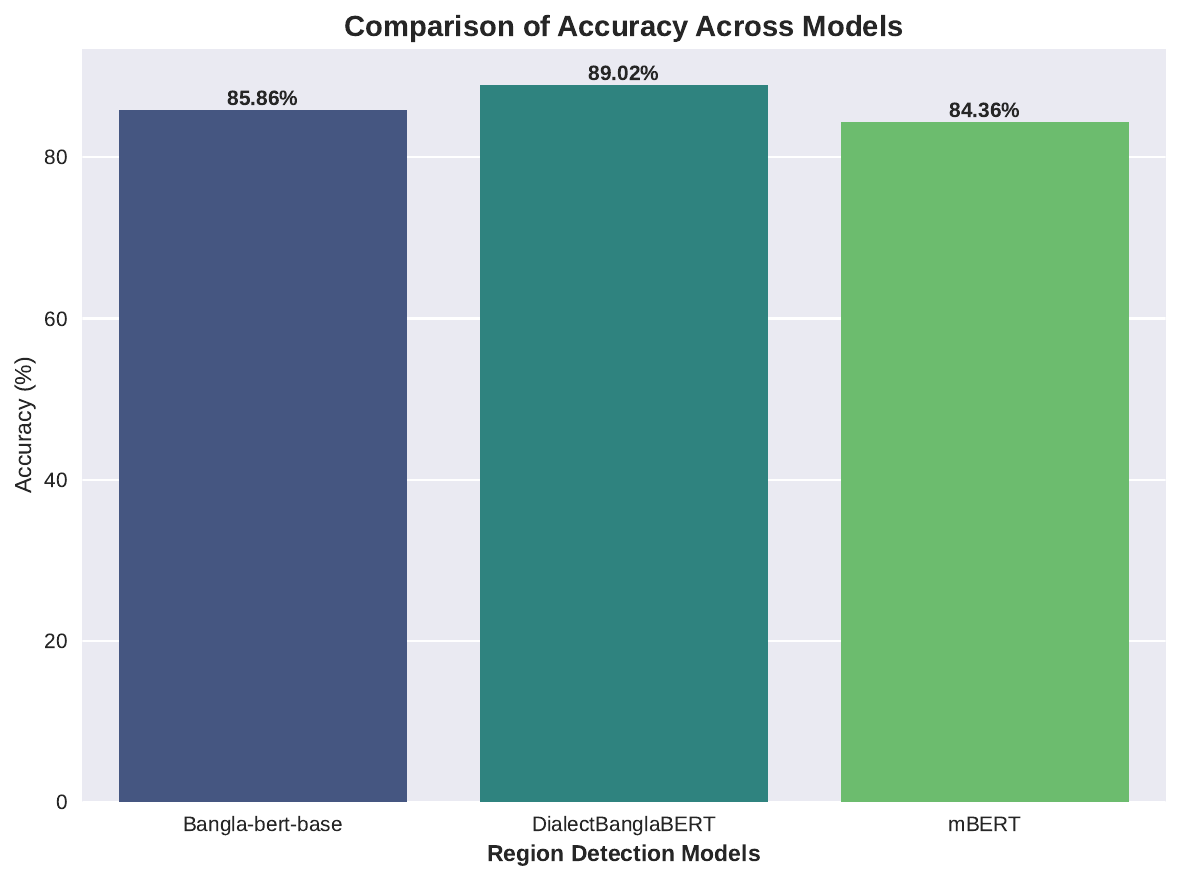}
        \caption{Accuracy Comparison}
       
    \end{subfigure}
    \hfill
    \begin{subfigure}{0.32\textwidth}
        \includegraphics[width=\linewidth]{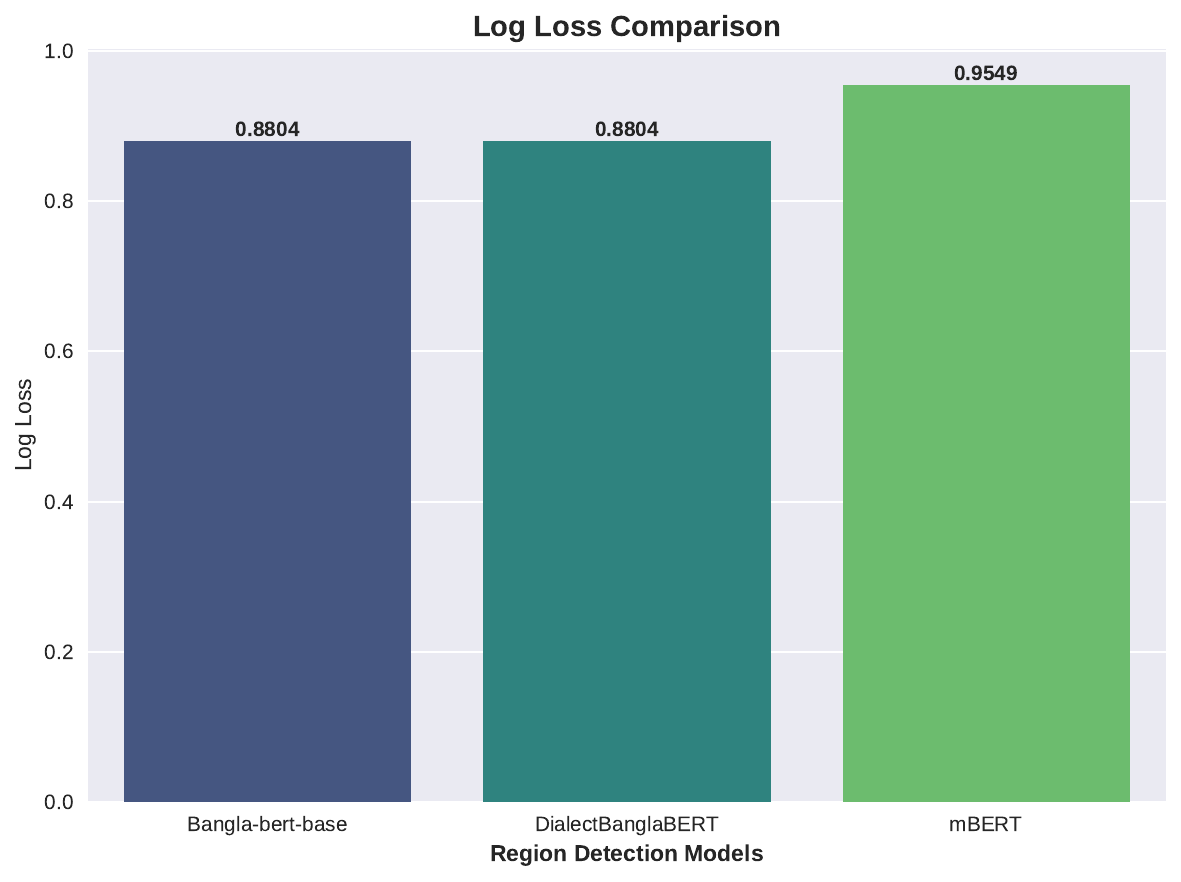}
        \caption{Weighted F1 Score Comparison} 
        
    \end{subfigure}
    \hfill
    \begin{subfigure}{0.32\textwidth}
        \includegraphics[width=\linewidth]{ComparisonLogloss.pdf}
        \caption{Log Loss Comparison}
      
    \end{subfigure}

    \caption{Performance comparison of models across three evaluation metrics: Accuracy, Weighted F1 Score, and Log Loss. Each subplot illustrates the model-wise performance under the respective metric.}
\label{com} 
\end{figure*}
\begin{figure}
\centering
\includegraphics[width=1.0\linewidth]{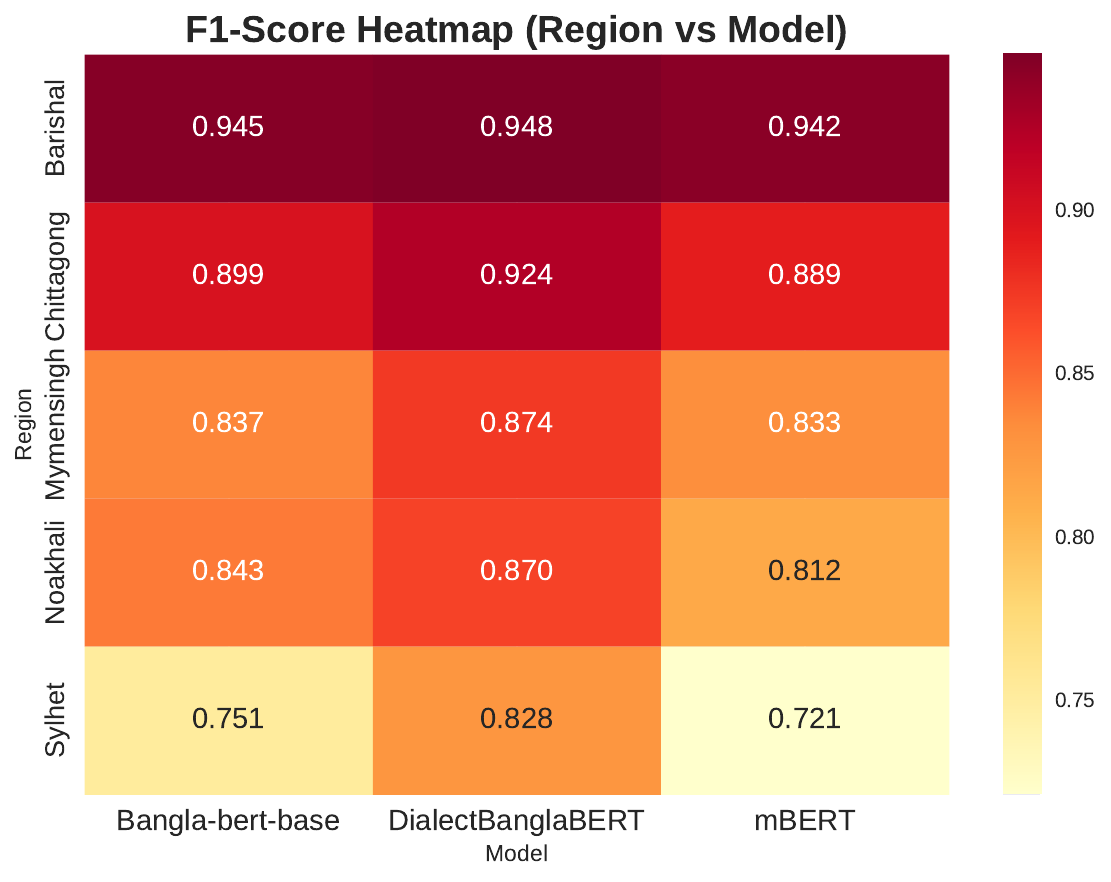}
\caption{F1-Score heatmap showing region-wise performance of three models: mBERT, Bangla BERT base, and DialectBanglaBERT across five Bangla dialect regions.}\label{F1Heatmap}
\end{figure}
\begin{figure}
\centering
\includegraphics[width=1.0\linewidth]{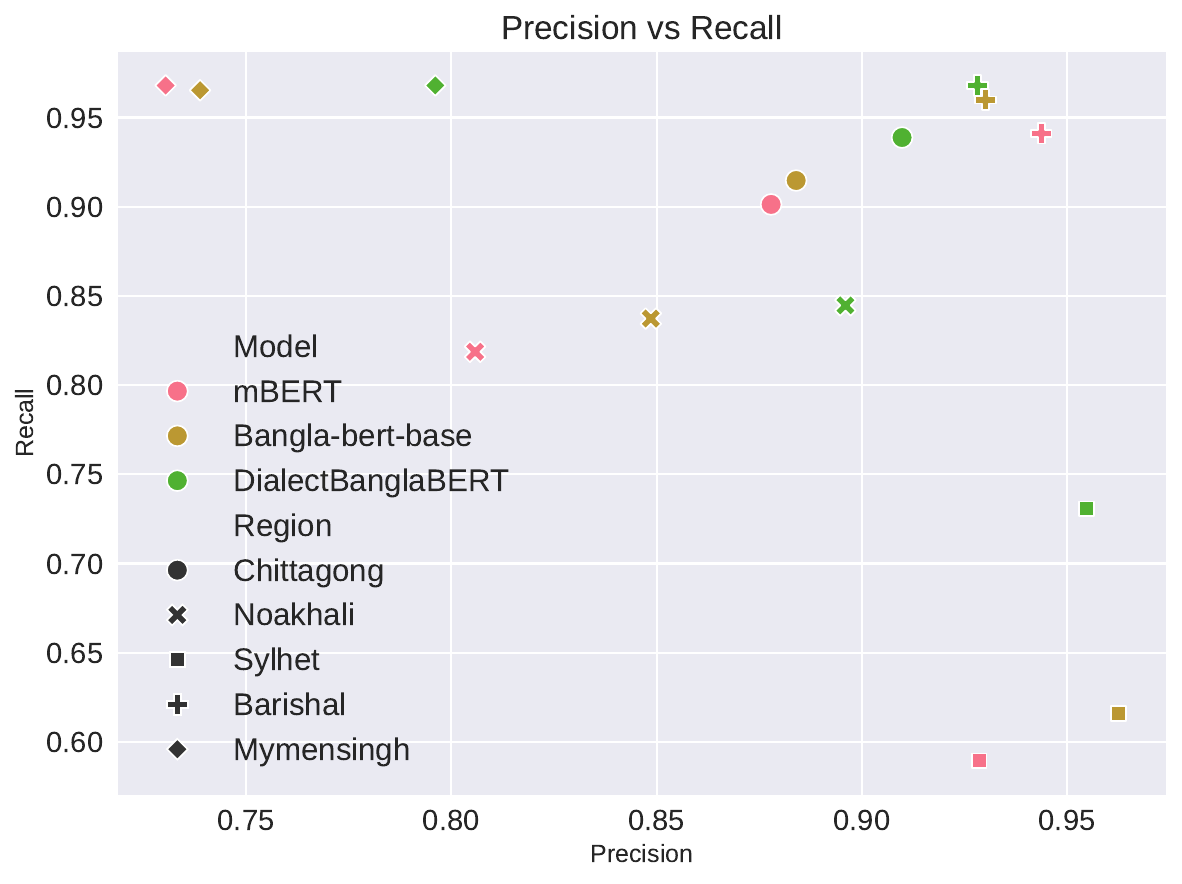}
\caption{Precision vs. Recall plot for three models: mBERT, Bangla BERT base, and DialectBanglaBERT evaluated across five Bangla dialect regions. Each point represents a model and region pair, highlighting the trade-off between precision and recall.}\label{Comparisonprevsrecall}
\end{figure}
Table \ref{table9} presents the performance comparison of three translation models: mT5, BanglaT5, and the proposed DialectBanglaT5, across five Bangla regional dialects: Chittagong, Noakhali, Sylhet, Barishal, and Mymensingh, using CER, WER, BLEU, and METEOR as evaluation metrics. Overall, DialectBanglaT5 consistently outperforms both baseline models across all dialects and metrics, demonstrating its effectiveness in capturing dialectal variations. For the Chittagong dialect, it achieves the lowest CER (0.2009) and WER (0.3225), along with the highest BLEU (46.91) and METEOR (0.6750), indicating a clear improvement over Bangl\-aT5. In the case of Noakhali, DialectBanglaT5 further enhances BLEU and METEOR scores to 49.85 and 0.6985, respectively, reflecting improved translation quality. For the Sylhet dialect, although mT5 performs well in CER and WER, DialectBanglaT5 outperforms all models with a BLEU score of 53.87 and METEOR of 0.7284. A similar pattern is observed in the Barishal dialect, where it achieves a BLEU of 55.94 and METEOR of 0.7528. The most notable results are seen in the Mymensingh dialect, where DialectBanglaT5 obtains the highest BLEU score of 71.93 and METEOR of 0.8503, while also maintaining the lowest CER (0.0791) and WER (0.1470). These findings highlight the significant advantage of integrating dialect-specific knowledge, with DialectBanglaT5 demonstrating superior performance and generalizability across all regional variations. 
\begin{table*}
\caption{Performance Overview of all region detection models}\label{table11}
  \begin{tabular}{lcclccc} 
    \toprule
 \textbf{Models} & \textbf{Accuracy} & \textbf{Log Loss} & \textbf{Region} & \textbf{Precision} & \textbf{Recall} & \textbf{F1-Score}  \\
    \midrule
    mBERT & 84.36\% & 0.9549 & Chittagong & 0.8779 & 0.9013  &  0.8895 \\
     &  &  & Noakhali  & 0.8058 & 0.8187  & 0.8122 \\
     &  &  & Sylhet  & 0.9286  & 0.5893 & 0.7210 \\
     &  &  & Barishal  & 0.9437 & 0.9412 & 0.9424 \\
     &  &  & Mymensingh  & 0.7304 & 0.9680 & 0.8326 \\
     \midrule
      Bangla-bert-base & 85.86\% & 0.8804 & Chittagong & 0.8840 & 0.9147  &  0.8991 \\
     &  &  & Noakhali  & 0.8486 & 0.8373 & 0.8430 \\
     &  &  & Sylhet  & 0.9625 & 0.6160 & 0.7512 \\
     &  &  & Barishal  & 0.9301  & 0.9599  & 0.9447 \\
     &  &  & Mymensingh  & 0.7388 & 0.9653 & 0.8370 \\ 
         \midrule
        DialectBanglaBERT  & 89.02\% & 0.8804 & Chittagong & 0.9098 & 0.9388  &  0.9241 \\
     &  &  & Noakhali  & 0.8960 & 0.8447 & 0.8696 \\
     &  &  & Sylhet  & 0.9547 & 0.7307 & 0.8278 \\
     &  &  & Barishal  & 0.9282  & 0.9679  & 0.9476 \\
     &  &  & Mymensingh  & 0.7961 & 0.9680 & 0.8736 \\
    \bottomrule
  \end{tabular}
\end{table*}

\begin{figure*}
\centering
\includegraphics[width=0.9\textwidth]{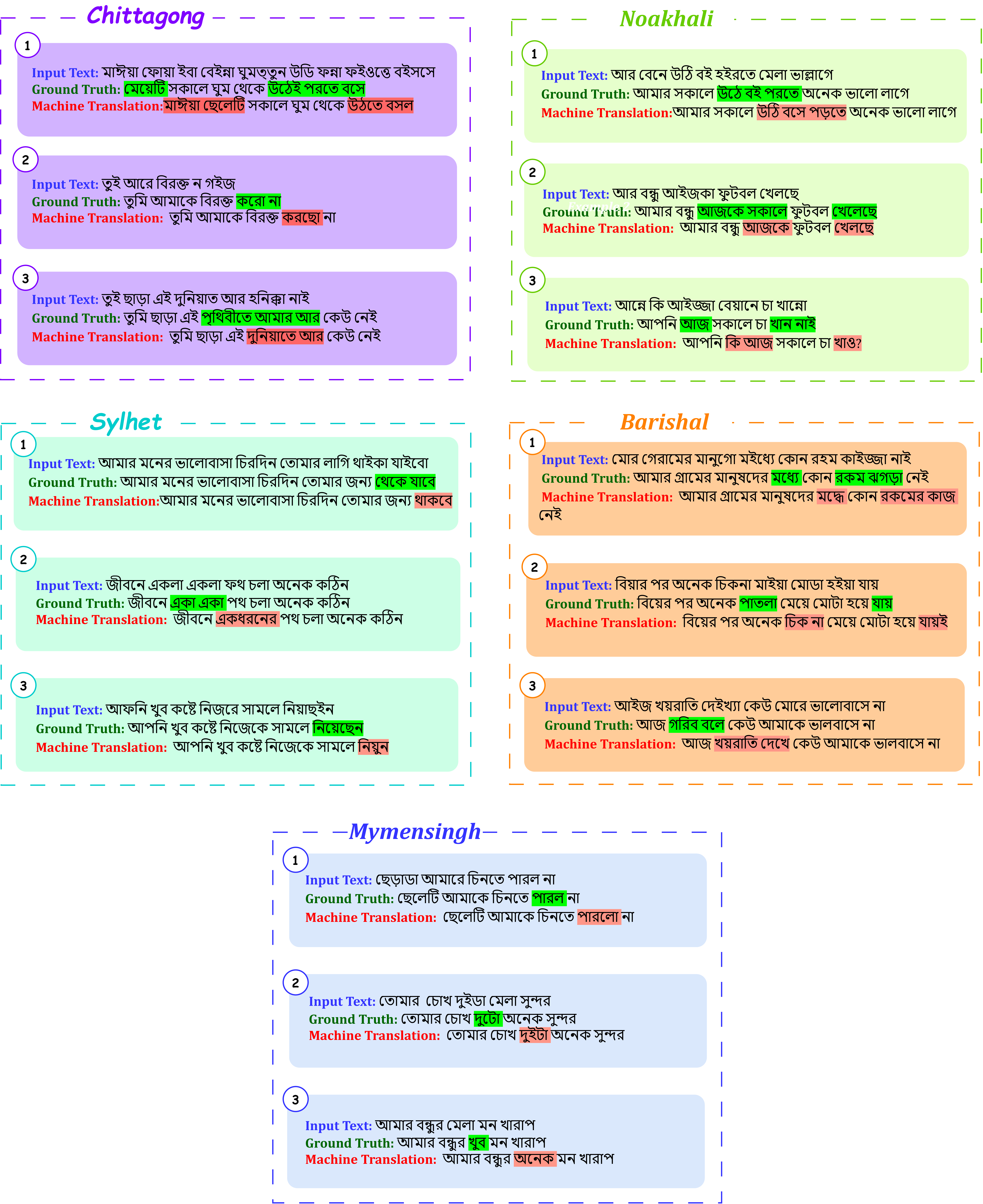}
\caption{Qualitative error analysis of machine-translated outputs from five Bangla regional dialects (Chittagong, Noakhali, Sylhet, Barishal, and Mymensingh) into standard Bangla. Each case highlights discrepancies between the ground truth and machine translations, revealing challenges in lexical choice, semantic preservation, and regional variation handling.}\label{errors}
\end{figure*}

Table \ref{table10} presents the ROUGE-1, ROUGE-2, and ROUGE-L scores (recall, precision, and F1) for mT5, BanglaT5, and DialectBanglaT5 across five Bangla regional dialects. DialectBanglaT5 consistently achieves the highest F1 scores across all dialects and ROUGE variants, demonstrating its effectiveness in handling dialectal nuances. For example, in the Chittagong and Noakhali dialects, it achieves ROUGE-1 F1 scores of 0.7325 and 0.7385, and ROUGE-L F1 scores of 0.7260 and 0.7365, respectively. Similar improvements are observed in Sylhet and Barishal, where DialectBanglaT5 outperforms the baselines in all metrics. The best results are seen in Mymensingh, with a ROUGE-1 and ROUGE-L F1 score of 0.8525 and ROUGE-2 F1 of 0.7248. These results highlight the model’s strong capability to produce accurate and fluent translations tailored to regional dialects. Figure~\ref{model_performance_comparison} shows a detailed performance comparison of translation models across five Bangla regional dialects using CER, WER, BLEU, and METEOR. Figure~\ref{DialectBanglaT5vsmt5} illustrates the percentage improvements of DialectBanglaT5 over mT5 across key evaluation metrics. Figure~\ref{DialectBanglaT5vsBanglat5} displays the metric-wise gains achieved by DialectBanglaT5 compared to BanglaT5, highlighting its effectiveness. Figure~\ref{Radar_Capped} presents a radar plot comparing BanglaT5, DialectBanglaT5, and mT5 in terms of CER, WER, BLEU, and METEOR across dialectal inputs.

Table \ref{table11} presents the performance comparison of three region detection models: mBERT, Bangla-bert-base, and proposed DialectBanglaBERT across five Bangla regional dialects. DialectBanglaBERT achieves the highest accuracy of 89.02 percent and consistently delivers better F1-scores across all regions. It performs strongly in Chittagong with an F1-score of 0.9241, in Noakhali with 0.8696, and in Mymensingh with 0.8736. Compared to mBERT and Bangla-bert-base, which show lower F1-scores in Sylhet and Mymensingh, DialectBanglaBERT provides a noticeable improvement, reaching 0.8278 in Sylhet. This highlights the effectiveness of dialect-specific adaptation in enhancing classification performance for regional variation in Bangla. The confusion matrices for these Three models are displayed in Figure \ref{figconf1},  Figure \ref{figconf2}, and Figure \ref{figconf3}. Moving on to the region-specific metrics, both models are evaluated on their precision, recall, and f1-score for five distinct regions: Chittagong, Noakhali, Sylhet, Barishal, and Mymensingh. In the Chittagong region, Bangla-bert-base shows a precision of 0.8840, recall of 0.9147, and an f1-score of 0.8991, indicating a balanced performance in correctly identifying instances of Chittagong. On the other hand, mBERT demonstrates lower a precision of 0.8779 and a lower recall of 0.9013 in the same region. For the Barishal region, mBERT exhibits a precision of 0.9437 and a recall of 0.9412, resulting in an f1-score of 0.9424. Bangla-bert-base, however, shows a slightly lower precision of 0.9301 and a higher recall of 0.9599, yielding a marginally higher f1-Score of 0.9447. Figure~\ref{F1Heatmap} shows a heatmap of F1-scores for mBERT, Bangla BERT base, and DialectBanglaBERT across five Bangla regional dialects, highlighting model-wise regional performance. Figure~\ref{com} presents a comparative analysis of model performance across Accuracy, Weighted F1 Score, and Log Loss, with separate subplots for each metric. Figure~\ref{Comparisonprevsrecall} illustrates the precision-recall trade-offs for mBERT, Bangla BERT base, and DialectBanglaBERT across all dialect regions, with each point representing a specific model-region pair.

\subsection{Performance Comparison}
The error analysis presented in Figure \ref{errors} offers a comprehensive view of the challenges encountered by the machine translation model when handling various Bangla regional dialects, including Chittagong, Noakhali, Sylhet, Barishal, and Mymensingh. Across these dialects, we identify common translation errors related to gender confusion, tense mismatch, lexical misunderstanding, and syntactic distortions.
In the Chittagong dialect, errors frequently stem from incorrect gender representation, tense inconsistency, and misinterpretation of culturally grounded expressions. For instance, the word “{\bng maIIya}'' (girl) is mistranslated as “ {\bng echeliT}'' (boy), introducing a clear gender mismatch. Another example reveals tense confusion, where “{\bng ibrk/t n g{I}j}'' (don’t bother me) is incorrectly rendered in the progressive tense. Moreover, emotionally charged terms like “{\bng Hink/ka,}'' which imply someone deeply personal, are flattened into generic expressions like “{\bng Aar ekU en{I},}'' stripping the sentence of its intended emotional nuance.
In the Noakhali dialect, the model struggles with verb form selection and negative constructions. A phrase like “{\bng UiTh b{I} H{I}ret}'' (wake up and study) is mistranslated as “{\bng UiTh bes pDet,}'' changing the meaning. Similarly, “{\bng ekhlech}'' (is playing) is used instead of the correct past form “{\bng ekhelech,}'' and the negative sentence “{\bng ca khaen/na}'' (did not drink tea) is turned into an interrogative “{\bng ca kha{O}?}'', completely altering the structure and meaning.
The Sylhet dialect reveals phonetic confusion, grammatical errors, and incorrect semantic interpretations. For example, “{\bng tha{I}ka Ja{I}eba}'' (will stay) is translated as “{\bng thakeb,}'' failing to capture the nuance of intent. Another sentence, “{\bng Ekla Ekla phth,}'' which refers to walking a lonely path, is misinterpreted as “{\bng Ekdhrenr pth}'' (a type of path), shifting the intended emotional and semantic tone. Additionally, the sentence “{\bng injer samel inyach{I}n}'' is grammatically distorted in translation, using an imperative form rather than the intended past perfect.
For the Barishal dialect, translation errors are mainly lexical and contextual. The phrase “{\bng manuega m{I}edhY}'' (among people) is incorrectly translated as “{\bng manuShedr med/dh,}'' which is close but leads to a completely different ending — “{\bng kaj en{I}}'' (no work) instead of “{\bng jhgDa en{I}}'' (no conflict). The dialectal adjective “{\bng ickna}'' (thin) is fragmented in translation, and socially contextual words like “{\bng khyrait}'' (used ironically to mean poor) are translated literally as “{\bng khJrait edekh,}'' missing the social undertone embedded in the original.
In the Mymensingh dialect, errors are relatively minor but still notable. The phrase “{\bng icnet parl na}'' (could not recognize) is slightly altered to “{\bng parela na,}'' introducing unnecessary formality. “{\bng ecakh du{I}Da}'' is rendered as “{\bng ecakh du{I}Ta,}'' which is grammatically correct but less standard than “{\bng ecakh dueTa}''. Lastly, “{\bng emla}'' (a dialectical intensifier meaning “very”) is translated as “{\bng Aenk,}'' which, while accurate, fails to preserve the intensity of the original sentiment.
The observed translation errors likely stem from limited dialectal representation in the training data, leading to poor generalization across regional linguistic variations. Additionally, dialect-specific vocabulary, phonetic irregularities, and non-standard grammar patterns challenge standard models trained on formal Bangla. The lack of contextual and cultural grounding further contributes to semantic mismatches and incorrect interpretations.

\section{Real-world Applicability and Deployment Considerations}

Deploying DialectBanglaT5 and DialectBanglaBERT in practical language processing systems involves important considerations related to accuracy, efficiency, and scalability, especially in resource-constrained environments. Since DialectBanglaT5 is based on a transformer architecture and handles inputs from diverse dialectal sources, it may face inference latency and memory overhead issues when applied to real-time scenarios such as mobile translation tools or regional news platforms. To address these challenges, techniques like model quantization, pruning, and knowledge distillation can help reduce the computational load while maintaining performance.

Scalability is essential for deployment across regions with significant dialectal variation. The proposed models can be integrated into educational technologies, regional language accessibility tools, and administrative translation services, all of which require accurate dialect handling. Additionally, embedding these models into voice assistants and real-time captioning systems can improve inclusivity for underrepresented dialects.

As dialects continue to evolve, ensuring adaptability is critical. Incorporating continual learning strategies such as online updates based on new dialectal data or active learning using user interaction can help maintain robustness over time. Ethical concerns must also be addressed, since misinterpretation or misclassification can lead to misunderstandings or reinforce cultural biases. Integrating a human-in-the-loop validation layer can reduce these risks, particularly for sensitive content.

By optimizing efficiency, enhancing adaptability, and incorporating responsible oversight, DialectBanglaT5 and DialectBanglaBERT provide a practical foundation for building dialect-aware Bangla NLP systems with real-world impact.
\section{Future Research Directions} \label{Futurelab}
While our work lays a strong foundation for regional Bangla dialect translation and region detection, several promising research directions can further expand its impact.
A key priority is improving the handling of slang and informal registers, which often carry rich cultural and emotional meanings. Expressions like {\bng ``samar epha''} from the Barishal dialect or {\bng ``egalaemr put''} from Mymensingh highlight the challenges of interpreting such language in a culturally appropriate manner. We plan to leverage large language models (LLMs) like GPT-4 or Claude 3.7 through in-context learning and few-shot prompting to better normalize and translate these expressions into standard Bangla. We also aim to integrate emotion recognition into the translation process. Beyond basic sentiment labels, detecting emotions such as anger, joy, fear, or sarcasm can provide a more nuanced understanding of regional speech. Emotion-aware translation can be achieved through multi-task learning or LLM prompting strategies tailored to capture affective intent, enhancing the emotional fidelity of translated outputs. Another direction involves cross-regional dialect translation, which remains largely unexplored. Translating between dialects such as Chittagong and Sylheti requires understanding deep linguistic variations. We aim to develop intermediate dialect embeddings or adapter layers to enable zero-shot or few-shot cross-dialect translation, supported by transfer learning techniques that capture phonological and syntactic differences.
The advent of instruction-tuned LLMs presents new opportunities for solving dialect tasks through prompt engineering. We plan to experiment with zero-shot and few-shot prompting, including chain-of-thought strategies to guide reasoning in dialect classification and generation. Soft prompt tuning and instruction fine-tuning will also be explored to adapt LLMs to dialect-specific contexts with minimal labeled data. To overcome data scarcity, we will employ LLM-based data augmentation. Techniques such as generating dialectal paraphrases from standard Bangla, creating slang-rich sentences, and producing emotion-labeled corpora will enrich the training set. Back-translation using fine-tuned LLMs will help simulate diverse dialect variants while maintaining linguistic authenticity. We also intend to translate our research into a deployable system. A mobile or web application could allow users to input speech or text in a dialect and receive its translation along with predicted region and emotional tone. Using efficient models like quantized LLMs, we will explore deployment in low-resource environments for broader accessibility.
Finally, recognizing the limitations of metrics like BLEU and ROUGE in dialectal contexts, we plan to develop human-centered evaluation protocols. These will include intelligibility testing, cultural sensitivity assessments, and expert reviews to ensure our models are not only technically sound but also socially responsible and linguistically accurate.

\section{Conclusion} \label{Conclusionlab}

Our research has made considerable advances in the translation of Bangla regional dialects into the standard Bangla language by introducing two dialect-aware models: DialectBanglaT5 for translation and DialectBanglaBERT for region detection. DialectBanglaT5 consistently surpasses baseline models across five major dialects: Chittagong, Noakhali, Sylhet, Barishal, and Mymensingh. The model achieves its best performance in the Mymensingh dialect, with the highest BLEU score of 71.93 and METEOR score of 0.8503, along with the lowest CER and WER. In contrast, the Chittagong dialect presents the greatest challenge, showing the lowest BLEU and METEOR scores, though still significantly improved compared to existing models. DialectBanglaBERT also achieves the highest overall accuracy of 89.02 percent in region classification. In the Chittagong region, it obtains an F1-score of 0.9241, reflecting its strong ability to capture regional linguistic cues. Across all dialects, it outperforms mBERT and Bangla-bert-base in both accuracy and F1-scores. These findings highlight the importance of incorporating dialect-specific knowledge into model design. Our work provides a valuable foundation for developing more inclusive, robust, and accurate NLP tools for low-resource and linguistically diverse communities.

\subsection*{CRediT authorship contribution statement}
\textbf{Fatema Tuj Johora Faria:} Conceptualization, Data curation, Methodology, Investigation, Software, Writing - Original Draft. \textbf{Mukaffi Bin Moin:} Conceptualization, Data curation,  Methodology, Investigation, Software, Writing - Original Draft. \textbf{Ahmed Al Wase:} Data Curation, Validation,  Software, Writing - Review \& Editing. \textbf{Mehidi Ahmmed:} Data Curation, Validation, Resources. \textbf{Md. Rabius Sani:} Data Curation, Validation, Resources. \textbf{Tashreef Muhammad:} Formal analysis, Writing - Review \& Editing, Supervision.

\subsection*{Funding}
This research was conducted without the involvement of any external funds or benefits. No external parties are involved with the conducted research.

\subsection*{Conflict of Interest}
The authors declare that none of them have any conflict of interest in the research presented in the manuscript. Thus, none of the authors have any financial or personal relationship that could potentially influence the research outcome.

\subsection*{Ethics Approval}
Not Applicable.

\subsection*{Consent to Participate}
The research was conducted with the help of 19 volunteers who have helped the authors in the events relating to the translation. All the participants were adults within the age group of 22 - 35 and they all agreed to participate willingly without any benefit and such with additional identity non-disclosure agreements.

\subsection*{Consent to Publishing}
All the participants have waived any rights to claim to the research work. All the authors have agreed to publish the research without any conflict.

\subsection*{Availability of Data and Material}
Data is provided within the manuscript cited in Availability and Usage Section \ref{Availability} which is publicly available through \href{https://data.mendeley.com/datasets/bj5jgk878b}{Mendeley Data}\footnote{\href{https://data.mendeley.com/datasets/bj5jgk878b}{https://data.mendeley.com/datasets/bj5jgk878b}} online. There are not any other materials associated with this research.

\subsection*{Code Availability}
The codes and associated coding files for this work are publicly available online on the \href{https://github.com/Mukaffi28/Vashantor-A-Large-scale-Multilingual-Benchmark-Dataset}{GitHub}\footnote{\href{https://github.com/Mukaffi28/Vashantor-A-Large-scale-Multilingual-Benchmark-Dataset}{https://github.com/Mukaffi28/Vashantor-A-Large-scale-Multilingual-Benchmark-Dataset}} platform. The codes are publicly accessible.

\vspace{2mm}

\bibliography{Reference}

\end{document}